\documentclass{article}

\usepackage{microtype}
\usepackage{graphicx}
\usepackage{subcaption} 
\usepackage[export]{adjustbox}
\usepackage{booktabs} 

\usepackage{hyperref}

\usepackage[table]{xcolor}

\usepackage[accepted]{icml2024}

\usepackage{amsmath}
\usepackage{amssymb}
\usepackage{mathtools}
\usepackage{amsthm}
\usepackage{multirow}

\usepackage[capitalize,noabbrev]{cleveref}

\theoremstyle{plain}

\theoremstyle{definition}

\theoremstyle{remark}

\usepackage[textsize=tiny]{todonotes}
\usepackage{svg}
\usepackage{float}
\usepackage{subcaption}
\usepackage{xcolor}
\usepackage{verbatim}

\newcommand{\ql}[1]{\textcolor{black}{#1}}
\newcommand{\wzl}[1]{\textcolor{black}{#1}}

\icmltitlerunning{VinT-6D: A Large-Scale Object-in-hand Dataset from Vision, Touch and Proprioception}

\begin{document}

\twocolumn[
\icmltitle{VinT-6D: A Large-Scale Object-in-hand Dataset\\ from Vision, Touch and Proprioception}

\icmlsetsymbol{equal}{*}

\begin{icmlauthorlist}
\icmlauthor{Zhaoliang Wan}{sysu,tencent}
\icmlauthor{Yonggen Ling}{tencent}
\icmlauthor{Senlin Yi}{sysu}
\icmlauthor{Lu Qi}{merced}
\icmlauthor{Wangwei Lee}{tencent}
\icmlauthor{Minglei Lu}{tencent}
\icmlauthor{Sicheng Yang}{tencent}
\icmlauthor{Xiao Teng}{tencent}
\icmlauthor{Peng Lu}{tencent}
\icmlauthor{Xu Yang}{casia}
\icmlauthor{Ming-Hsuan Yang}{merced}
\icmlauthor{Hui Cheng}{sysu}

\end{icmlauthorlist}

\icmlaffiliation{sysu}{School of Computer Science and Engineering, Sun Yat-sen University, Guangzhou, China}
\icmlaffiliation{tencent}{Robotics X, Tencnet, Shenzhen, China}
\icmlaffiliation{merced}{The University of California, Merced, Merced, the U.S.}
\icmlaffiliation{casia}{Chinese Academic of Sciences, Automation Institute, Beijing, China}
\icmlcorrespondingauthor{Yonggen Ling}{rolandling@tencent.com}
\icmlcorrespondingauthor{Hui Cheng}{chengh9@mail.sysu.edu.cn}

\icmlkeywords{Machine Learning, ICML}


\icmlkeywords{Perception for Robotic Grasping and Manipulation, Touch Sensing, Embodeid AI, Pose Estimation}

\vskip 0.3in
]

\printAffiliationsAndNotice{Zhaoliang Wan conducted this research during his internship at Robotics X, Tencent.}

\begin{abstract}
\ql{This paper addresses the scarcity of large-scale datasets for accurate object-in-hand pose estimation, which is crucial for robotic in-hand manipulation within the ``Perception-Planning-Control" paradigm. Specifically, we introduce VinT-6D, the first extensive multi-modal dataset integrating vision, touch, and proprioception, to enhance robotic manipulation. VinT-6D comprises 2 million VinT-Sim and 0.1 million VinT-Real splits, collected via simulations in MuJoCo and Blender and a custom-designed real-world platform. This dataset is tailored for robotic hands, offering models with whole-hand tactile perception and high-quality, well-aligned data. To the best of our knowledge, the VinT-Real is the largest considering the collection difficulties in the real-world environment so that it can bridge the gap of simulation to real compared to the previous works. Built upon VinT-6D, we present a benchmark method that shows significant improvements in performance by fusing multi-modal information. The project is available at \href{https://VinT-6D.github.io/}{https://VinT-6D.github.io/}.
}


\end{abstract}

\begin{figure}[t!]
\vskip 0.2in
\begin{center}
\centerline{\includegraphics[width=1\columnwidth]{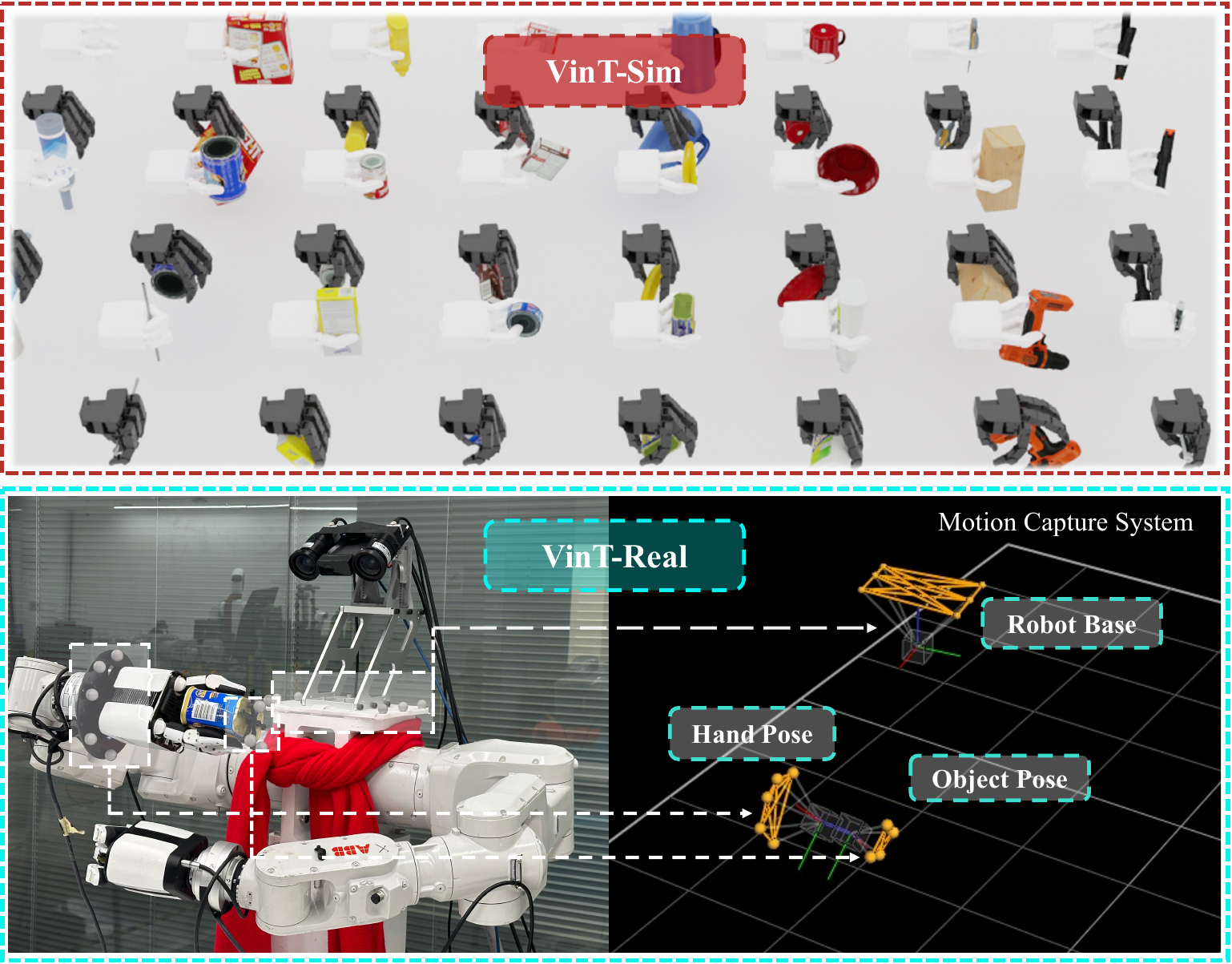}}
\vskip -0.1in
\caption{\wzl{\textbf{Large-scale object-in-hand dataset VinT-6D comprising synthesized and real-world splits naming VinT-Sim and VinT-Real.} VinT-Sim aims to generate realistic data across vision, touch, and proprioception. VinT-Real is gathered through a precisely calibrated and aligned multi-modal robot platform, where a motion capture system obtains accurate object and hand poses.} }
\label{fig-0}
\end{center}
\vskip -0.4in
\end{figure}

\section{Introduction}
\label{submission}

\ql{Estimating 6D object-in-hand pose~\cite{wang2019densefusion, dikhale2022visuotactile} is a crucial area in both computer vision and robotics, that benefits numerous applications such as dexterous manipulation~\cite{kelestemur2022tactile, qi2023high, suresh2023neural} and grasping~\cite{liang2020hand}. 
Specifically, utilizing multi-modal signals, including vision, touch, and proprioception, is emerging as a significant research trend due to their unique characteristics~\cite{liang2020hand, cui2021hand, rostel2022learning, kelestemur2022tactile, lin2023tactile}. 
These approaches align closely with human biological mechanisms, where the coordination of eyes, muscles, and joints is central to manipulating objects held in the hands.}

\begin{table*}[!t]
\caption{\textbf{Datasets for Object-in-Hand Pose Estimation.} Such datasets are rare and often constrained by size, cost, hardware configuration (including hand types and tactile sensor varieties), and ground truth quality. The symbol $\circ$ indicates datasets not to be publicly released or slated for future release. \checkmark and - indicate whether has some modality or location of tactile sensors.}
\label{tab:datasets}
\vskip 0.15in
\vspace{-4mm}
\centering
\footnotesize
\begin{tabular}{lccccccccccc}
\toprule
\multirow{2}*{\textbf{Dataset}} & \multirow{2}*{\textbf{Num of Fingers}}  & \multicolumn{3}{c}
{\textbf{Available Modalities}} && \multicolumn{3}{c}{\textbf{Tactile Distribution}} && \multicolumn{2}{c}{\textbf{Data Scale}} \\
\cmidrule(lr){3-5} \cmidrule(lr){7-9} \cmidrule(lr){11-12}
&& Vision  & Touch & Prop. && Tip & Pulp & Palm && Sim & Real \\
\midrule
Hand-Object~\cite{wen2020robust} & 2  & \checkmark & - & \checkmark && - & - & - && 12 K & 1 K \\
VITA~\cite{dikhale2022visuotactile} & 4 & $\circ$ & $\circ$ & $\circ$ && \checkmark & \checkmark & - && 0.2 M & - \\
Fast-Grasp'D~\cite{turpin2023fast} & 3, 4, 5 & $\circ$ & - & - && - & - & - && 1 M & - \\
TEG-Track~\cite{liu2023enhancing} & 2 & \checkmark & \checkmark & \checkmark && \checkmark & - & - && - & 0.2 K \\
PoseFusion~\cite{tu2023posefusion} & 5 & \checkmark & \checkmark & \checkmark && \checkmark & - & - && - & $\circ$ \\
FeelSight~\cite{suresh2023neural} & 4 & $\circ$ & $\circ$ & $\circ$ && \checkmark & - & - && - & $\circ$ \\
\rowcolor{violet!10} VinT-6D (Ours) & 3, 4 & \checkmark & \checkmark & \checkmark && \checkmark & \checkmark & \checkmark && 2 M & 0.1 M \\
\bottomrule
\end{tabular}
\end{table*}

\ql{Despite significant progress in representation learning across three modalities, this field faces two major issues due to the existing low-quality real-world dataset. 
First, due to the scarcity of large-scale public data,  
one common practice is to resort to synthesized data for training, resulting in a substantial domain gap with real-world environments at test time. 
Second, as existing methods focus on synthesized data for two-finger grippers, they could not be applied to complex scenarios such as occlusions in three or four-finger grasping. 
Both aspects hinder the further deployment of 6D object-in-hand pose estimation in robotic applications.}


\ql{Collecting a high-quality dataset for 6D object-in-hand pose estimation is challenging. The usage of various tactile sensors and robotic hand configurations makes it hard to generalize findings across different robotic hands. Gathering real-world data is also tricky due to time constraints, sensor malfunctions, imprecise data, and the difficulty of aligning multiple modalities. To achieve high-quality data collection, it is essential to establish a robust integration of hardware and software that can address these challenges and ensure data integrity.}

\ql{We address the above-mentioned problems and build the large-scale multi-finger object-in-hand perception dataset VinT-6D that comprises synthesized and real-world splits naming VinT-Sim and VinT-Real (\cref{fig-0}). 
To the best of our knowledge, our dataset performs better than existing datasets~\cite{dikhale2022visuotactile, turpin2023fast} in the aspect of both numbers and quality (\cref{tab:datasets}).} \wzl{Specifically, VinT-Sim focuses on bridging the sim2real gap in visual, tactile, and proprioceptive sensing by developing a consistent representation of tactile readings and accurately simulating taxel distributions, as detailed in the touch signal modeling section (\cref{sec-sim-touch}). The process involves loading the robotic hand and object into MuJoCo \cite{todorov2012mujoco} for iterative stable grasping simulations (\cref{sec-sim-interactions}), followed by rendering in Blender \cite{blender} for multi-view, realistic vision data (\cref{sec-sim-vision}); }

\wzl{VinT-Real captures multi-modal data through our custom-developed robotic platform, featuring an integration of visual, tactile, proprioceptive sensors and a motion capture system. We meticulously customize marker assemblies on fixed parts of each object, ensuring sub-millimeter level accuracy in object pose acquisition without hindering the object's functionality (\cref{sec-real-alignment}). The platform's calibrated sensors precisely align visual and tactile data, allowing the robot to emulate toddler-like exploratory movements in various grasp poses, including shifts and rotations within the workspace. 
 }

\ql{Based on the VinT-6D dataset, we propose a fundamental baseline VinT-Net for accurate object-in-hand pose estimation with synergizing vision, touch, and proprioception signals. 
Our key design is \wzl{the integration of touch and proprioception with vision, synergistically merging diverse and complementary sensory inputs}. This design could benefit \wzl{for object-in-hand pose estimation, particularly when vision is obscured by the robot's hand occlusions. Embedded touch sensors in fingers and palms and proprioceptive data from hands and arms offer valuable supplementary information}. 
Extensive experiments show the effectiveness of our method compared with the other works.}

\section{Related Work}

\subsection{Object Pose Estimation from Vision}

In computer vision, generic object pose estimation datasets such as LineMod \cite{hinterstoisser2011multimodal} and YCB-Video \cite{calli2017yale} provide valuable data, including color and depth images, segmentation labels, and object poses, 
facilitating significant advances in object pose estimation \cite{wang2019densefusion, peng2019pvnet, li2019cdpn, he2020pvn3d, chen2020g2l, he2021ffb6d, Wang_2021_self6dpp, chen2022epro, lipson2022coupled}. 
However, existing datasets are less suitable for robotics applications demanding higher precision. 
Recently, a new dataset for robotic manipulation with millimeter-level pose accuracy was introduced \cite{tyree20226}, but it focuses on objects on tables, not addressing in-hand scenarios. The study by \cite{wen2020robust} is notable for using hand state estimation to improve pose accuracy and for providing a dataset for a two-fingered gripper. However, in-hand perception is challenging for multi-fingered hands due to vision occlusions. This highlights the need for a dataset to benchmark vision-based multi-fingered object-in-hand poses.

\subsection{Object Pose Estimation from Touch}

\wzl{Tactile signals are crucial for object-in-hand pose estimation during manipulation, as shown in studies like \cite{liang2020hand, rostel2022learning, lin2023tracking, kelestemur2022tactile, lin2023tactile}, with most methods using touch and proprioception data to iteratively refine object pose models through filtering theory. However, these approaches, relying on sensitive tactile feedback and lacking vision's global perspective, often struggle with consistent performance, underlining the challenges of relying solely on tactile sensing. For a broader and more comprehensive review of robotic tactile perception, readers are directed to \cite{li2020review}.}

\subsection{Object Pose Estimation from Vision and Touch}
\wzl{Recent advancements in visual-tactile 6D object-in-hand pose estimation, initiated by the simulation dataset from \cite{dikhale2022visuotactile} and extended by \cite{li2023vihope} and \cite{rezazadeh2023hierarchical}, have shown promise. However, these studies, including simplistic simulations of vision and touch, reveal a significant domain gap and highlight the need for more realistic data synthesis and real-world collection.}
\wzl{Contributions like the small-scale dataset from \cite{tu2023posefusion} using a Shadow hand with BioTac sensors \cite{shadowhand}, \cite{liu2023enhancing}'s sequential visual-tactile tracking framework, and \cite{suresh2023neural}'s exploration with DIGIT sensors \cite{lambeta2020digit} on the AllegroHand \cite{simlab2016allegro}, demonstrate the growing interest in this field. Yet, these works have certain limitations in real-world applications, such as limited tactile data from fingertips only, reliance on two-finger grippers or external cameras to prevent occlusions, and stationary palm (comparison in \cref{tab:datasets}). These highlight current research gaps and the pressing need for a comprehensive and publicly available dataset that realistically represents multi-finger object-in-hand pose for robotic manipulation in real-world settings.}

\section{VinT-6D Dataset}

\wzl{We introduce the VinT-6D dataset, a comprehensive large-scale collection that combines two distinct data types: VinT-Sim and VinT-Real. VinT-Sim, designed to realistically synthesize and minimize the sim-to-real gap, contributes two million visual, tactile, and proprioceptive samples. Meanwhile, VinT-Real has 0.1 million high-quality, real-world data instances, all meticulously gathered from a precisely calibrated robotic system. The dataset features 25 household objects selected for their size, material, and utility diversity.}

\subsection{VinT-Sim}

\begin{figure}[t!]
\vskip 0.2in
\begin{center}
\centerline{\includegraphics[width=\columnwidth]{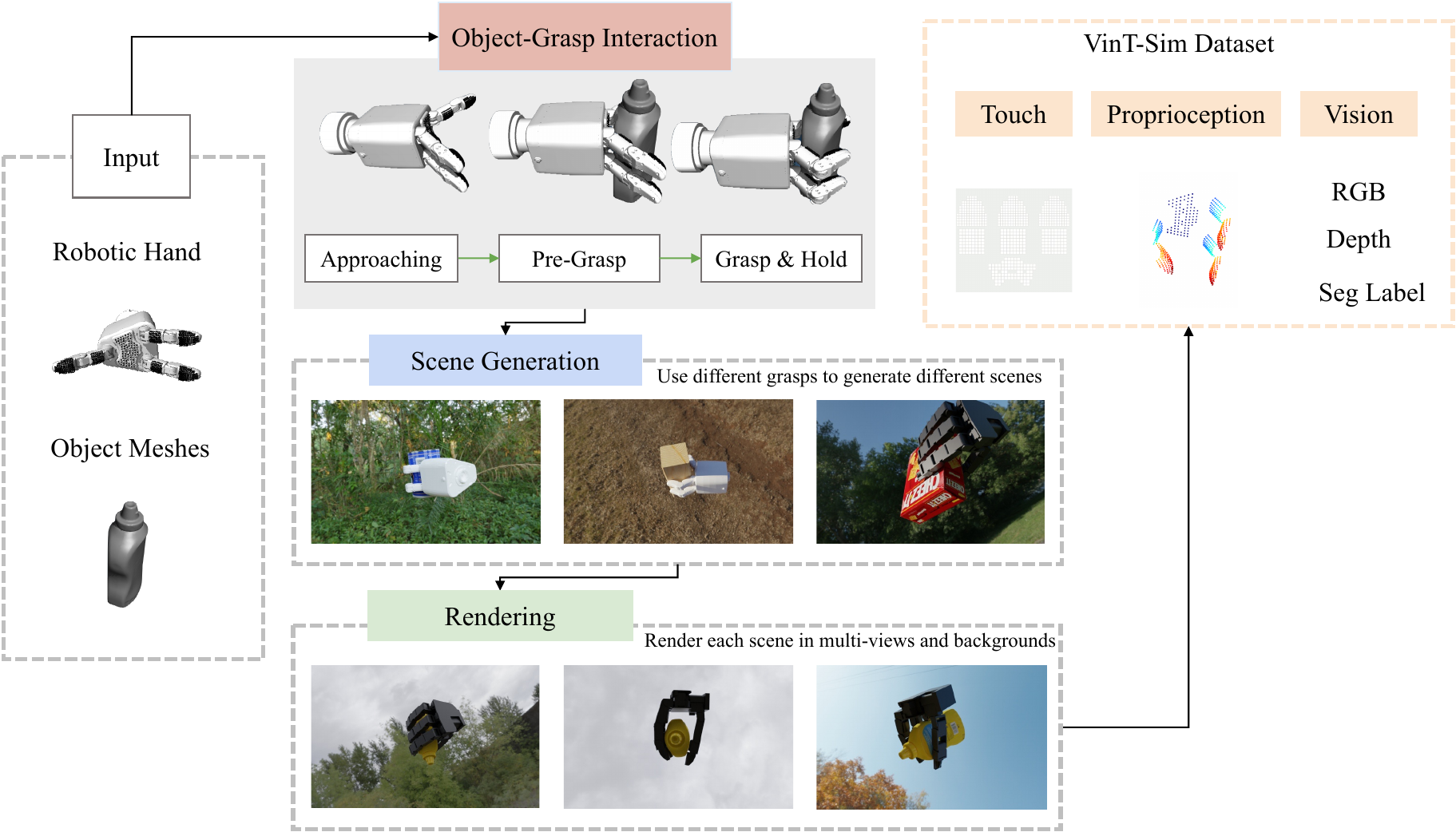}}
\vskip -0.1in
\caption{\textbf{VinT-Sim Dataset Generation Pipeline.} VinT-Sim requires a robotic hand as input, which can have either three or four fingers along with an object model. There are three components involved in this process: (1) Simulating whole-hand touch. (2) Generating tactile data and proprioception information through object-grasp interactions. (3) Rendering each object-grasp scene with various realistic backgrounds and capturing multiple views. }
\label{vint-sim-pipeline}
\end{center}
\vskip -0.4in
\end{figure}

\wzl{Collecting large-scale data for robotic in-hand perception presents considerable challenges due to the time-intensive process, sensor wear and tear, pose imprecision, and difficulties in aligning different modalities. Thus, the simulator plays a crucial role in synthesizing such extensive data, leading to our motivation to build the VinT-Sim split. A major challenge for such a collection is narrowing the gap between simulation and reality, approached from three perspectives: touch, proprioception, and vision. The process of generating datasets is illustrated in \cref{vint-sim-pipeline}.}

\subsubsection{Simulating Touch}
\label{sec-sim-touch}


\wzl{\textbf{Difficulties of simulating tactile modality.} Exploring tactile sensors, such as pressure and vision-based sensors, presents challenges for accurate and consistent representation in simulation with their unique characteristics. Concretely, the non-linear positive correlation to applied pressure for each taxel complicates direct simulation efforts, and the sensor performance degradation over time affects reading fidelity, necessitating dynamic simulation adjustments. Additionally, inherent performance differences between taxels due to manufacturing inconsistencies require individual calibration, making the direct simulation of tactile readings non-trivial.} 

\wzl{\textbf{Consistent representation of tactile readings.} We focus on the simulation of contact positions rather than mimicking each taxel's specific measurements. For both pressure and vision-based tactile sensors, well-calibrated contact positions can be robust to contact force and time variation. Our approach involves simulating each taxel as a force sensor with a non-linear response to the applied force. For each tactile sensor, we represent it as a cuboid cylinder topped with a blue hemisphere on the finger to generate touch signals and sensor reading upon contact and record the activated tactile sensor's position as the tactile data. For a detailed insight into the simulating touch sensor modeling and calibration process, please refer to \cref{app-touch-sim}.}

\begin{figure}[t!]
\begin{center}
\centerline{\includegraphics[width=\columnwidth]{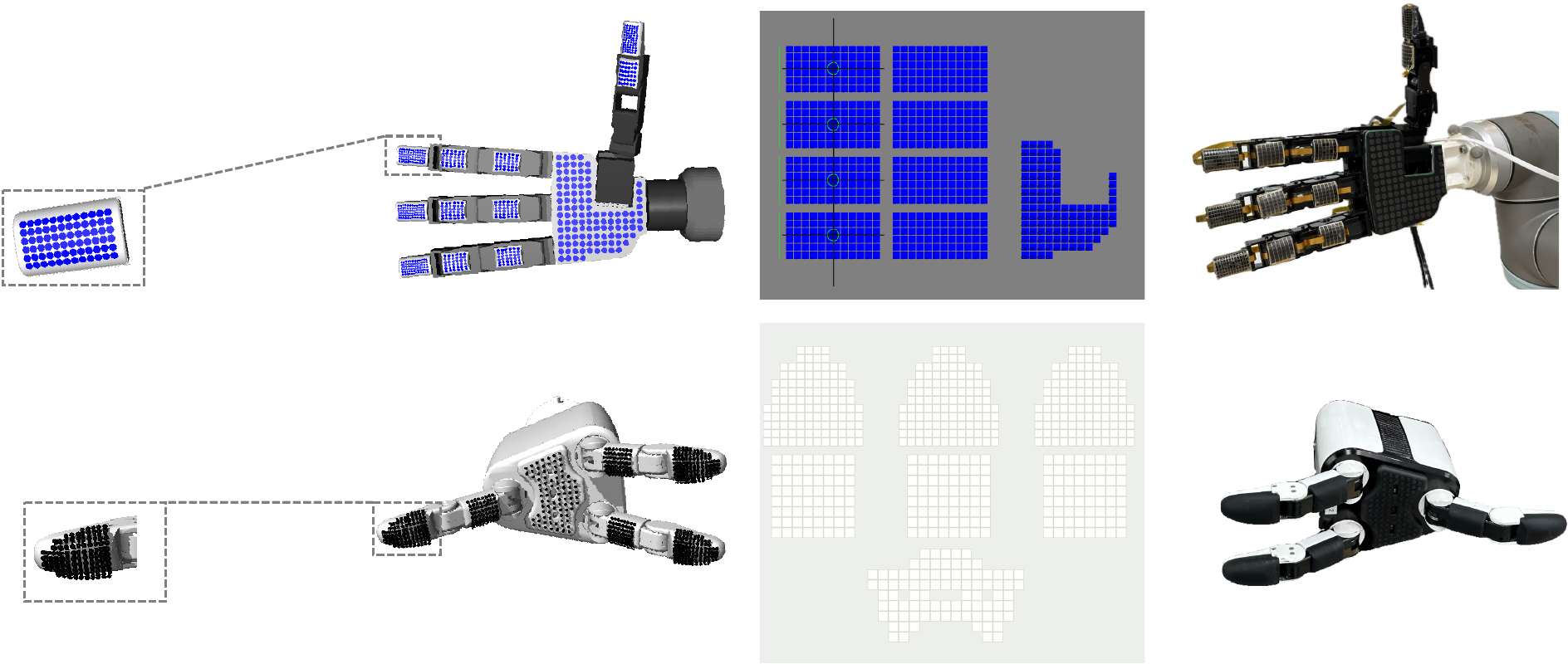}}
\vskip -0.1in
\caption{\textbf{Simulated and Real-world Robotic Hands with Whole-Hand Tactile Perception.} In VinT-6D, both the three-fingered Trx hand and the four-fingered Allegro hand are used to generate or collect datasets. These robotic hands are equipped with array-based tactile sensors covering the entire hand, with the simulated sensors distributed similarly to the real-world setup.}
\label{whole-hand-perception}
\end{center}
\vskip -0.5in
\end{figure}

\wzl{\textbf{Alignment to taxel distribution in real world.} For all tactile sensors on the whole hand, it is essential to carefully calibrate the robot's hand, tactile sensors, and their spatial relationship to accurately replicate tactile contact in simulations. For our real-world setup (refer to \cref{whole-hand-perception}), we equip two robotic hands (three-fingered Trx and four-fingered Allegro) with array-based tactile sensors. These sensors, with 620 and 679 taxels respectively, are piezoresistive, curved, and unlike flat arrays, have various taxel distributions due to being equipped on the surface of the whole hand. We meticulously scan each sensor's location relative to the hand's parts to ensure accurate replication of taxel distribution in the simulation. When the robotic hand holds an object, the activated tactile positions are recorded as a local touch point cloud ($D_{touch}$) relative to the palm frame, providing a robust and precise simulated tactile signal.}

\subsubsection{Simulating Object-Grasp Interaction}
\label{sec-sim-interactions}
Various object-grasp types and hand configurations influence the diversity of the dataset. Therefore, our simulation should include appropriate grasp types and account for different hand configuration

\textbf{Selected stable grasping.} Recent advances, such as \cite{turpin2023fast} and \cite{wan2023unidexgrasp++}, have significantly contributed to the generation of stable multi-finger grasps. However, these advancements have primarily focused on grasp stability and have not taken post-grasp task-specific manipulation into account. As a result, their practical application in real-world robotic manipulation might be limited. This leads us to ask: What form of interaction is required between a robotic hand and an object to enable grasping for manipulation? Our answer is selected stable grasping involves not only stable holding of the object but also readiness for manipulation. 
We develop a MuJoCo-based simulation for generating realistic object-grasp interactions. In this setup, we carefully fix the robotic hand and the object in predetermined poses within the simulation environment. To ensure a lifelike pre-grasp stance, we tangentially align the robotic hand's palm with a randomly selected point on the object's graspable surface. By employing inverse kinematics to move the finger joints, our system allows for tactile sensor contact with the object. Additionally, the proximity between the hand and the object follows a set ratio, further enhancing the simulation's realism.

\textbf{Replicating real-world scenarios.} Our approach to grasping objects is based on replicating real-world scenarios to achieve effective tactile engagement. By conducting numerous grasping processes for two hands with different configurations (three-fingered and four-fingered), we have been able to create a dataset of 125000 unique successful graspings. Our focus has been ensuring the grasp's stability and usefulness for future manipulative tasks. To achieve this, the hand executes a power grasp upon contact, lifting the object to a predetermined height for a physically realistic integration. We meticulously document the proprioception data, including the poses of the object ($P_{obj}$), hand ($P_{hand}$), and finger knuckles ($P_{knuckle}$), for subsequent vision rendering. The object-grasp setups are detailed examples, which can be found in \cref{app-obj-grasp}.

\subsubsection{Simulating Vision}
\label{sec-sim-vision}

Generating realistic visual data is essential for bridging the gap between simulation and real-world scenarios \cite{movshovitz2016useful}. While MuJoCo provides accurate physical simulations, its basic rendering capabilities limit the visual realism of the simulated environments. 

\textbf{Photo-realistic object-in-hand rendering.} To achieve this, we leverage Blender's advanced rendering features, including ray tracing, diverse shaders, and real-time viewport rendering through its Cycles engine. Based on recorded proprioception data, we can create photo-realistic object-in-hand images by importing object and hand models into Blender and setting their poses and finger knuckle positions. To enhance realism, we use various HDRI backgrounds for illumination and diverse scene backdrops, rendering each setup from multiple viewpoints within a hemispherical range. To ensure authenticity, we also carefully manage ray and glossy reflection bounces in reflective objects. As illustrated in \cref{app-sim-vision}, these techniques enable us to generate highly realistic visual data, improving domain adaptation.

\textbf{Real-world camera characteristics.} We use a multi-step process to enhance the realism of our simulated depth images. We start by rendering depth images in Blender and then apply post-processing techniques to simulate real-world conditions, specifically for those captured by a Kinect Azure TOF camera. We consider various factors such as distance, lighting, and surface specularity. To enhance realism, we introduce noise, create holes, and smooth the rendered depth images, following methodologies like those in \cite{tolgyessy2021evaluation}. By replicating the Kinect Azure sensor's unique noise characteristics, we aim to accurately recreate real-world conditions in our simulations. Finally, we convert the depth images to point clouds and align them with the local touch point clouds at the camera frame. This process is crucial for ensuring our depth images are as realistic and accurate as possible.

\subsection{VinT-Real}

To acquire accurate robotic in-hand perception data, it is crucial to establish a reliable hardware-software platform as well as sensor calibration and maintain precise data alignment. We detail our approach to ensure the quality and diversity of the collected data.

\subsubsection{\wzl{Modality Alignment}}
\label{sec-real-alignment}

\begin{figure}[t!]
\vskip 0.2in
\begin{center}
\centerline{\includegraphics[width=\columnwidth]{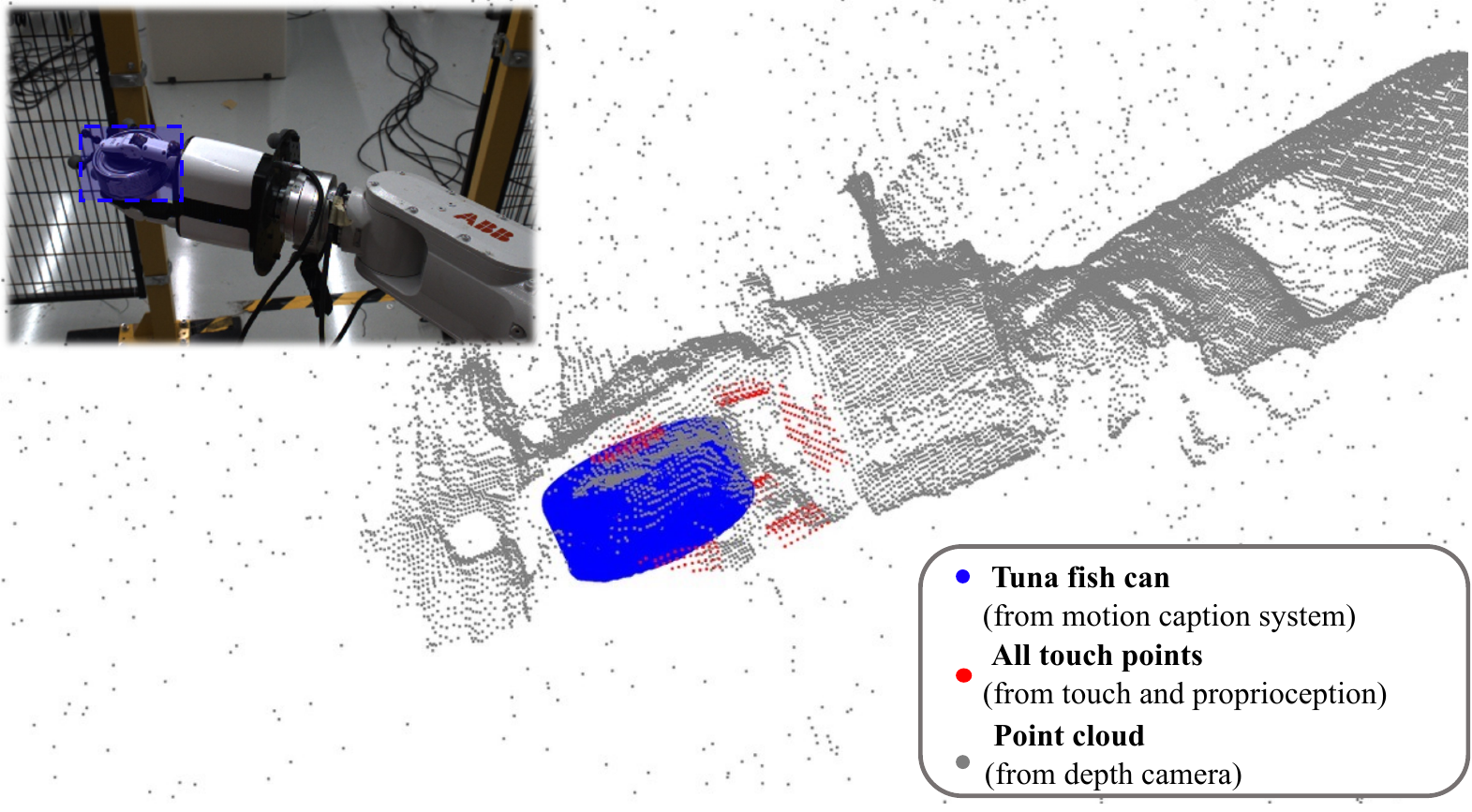}}
\vskip -0.1in
\caption{\textbf{Visualization of full tactile points aligned with vision in VinT-Real.} Gray points represent the point cloud from the depth camera, blue points depict the transformed model from the motion capture system, and red points indicate the full touch points of the hand.}
\label{vision-touch-align}
\end{center}
\vskip -0.4in
\end{figure}

\wzl{\textbf{Intra-modal Alignment.} VinT-Real needs to align depth images from an Azure Kinect TOF camera with RGB images from a stereo camera setup. We achieve this through precise camera calibration and distortion correction, resulting in well-aligned vision images. However, segmenting objects in hand can be a challenging task due to occlusions. To overcome this, our solution utilizes touch and proprioception as cues for the SAM segmentation model \cite{kirillov2023segment}. The SAM model is informed of activated touch points identified via robotic kinematics, which improves the segmentation process, as illustrated in \cref{seg-by-sam}. You can find more information on our real vision application in \cref{app-real-vision}.}

\textbf{Inter-modal Alignment.} We use touch positions to create a touch point cloud that is then translated into real-world applications through a calibrated robotic system. We calculate the hand pose relative to the robot's base using forward kinematics, starting from the finger knuckles to the hand frame. Our approach accurately maps activated touch sites onto the knuckles' positions to establish the touch point cloud. We ensure sub-millimeter accuracy object-in-hand pose (\cref{app-real-pose}) by augmenting the Trx-hand with a flange plate featuring markers, which are captured by a motion capture system. Our custom-designed fixtures equipped with markers for each object guarantee precise and consistent pose data when objects are held in hand. We make all fixture models available on our website. Our well-aligned multi-modal data is visualized in \cref{vision-touch-align}. Overall, our approach surpasses common ArUco tag-based methods \cite{calli2017yale, dikhale2022visuotactile, xu2023visual} and guarantees precise alignment with the vision system (\cref{app-real-vision}).

\begin{figure}[t!]
\vskip 0.2in
\begin{center}
\centerline{\includegraphics[width=\columnwidth]{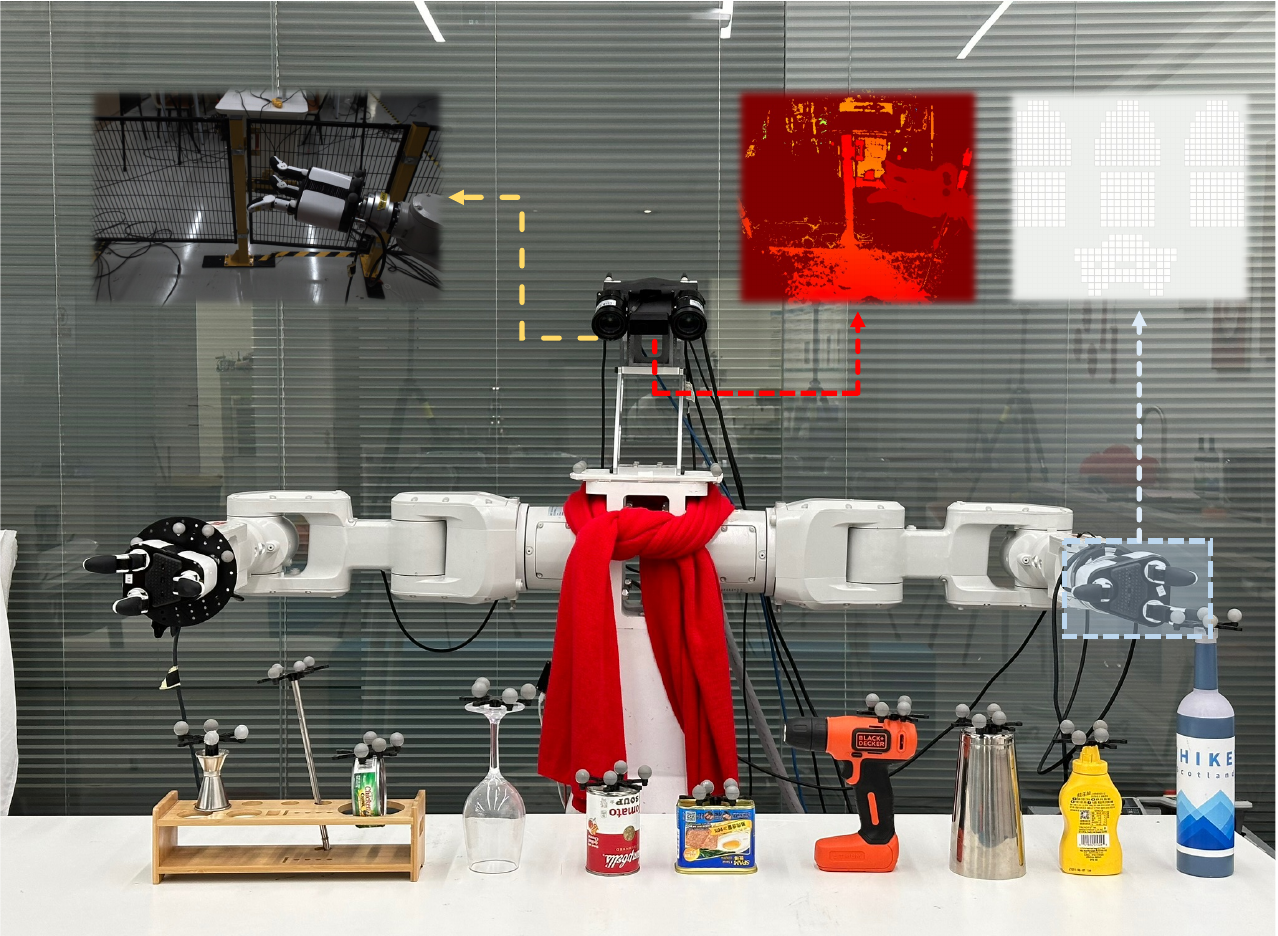}}
\vskip -0.1in
\caption{\textbf{Custom-Developed Robot Platform.} The objects selected in VinT-Real are neatly placed on the desk for easy access and manipulation.}
\label{bartender-robot}
\end{center}
\vskip -0.4in
\end{figure}

\subsection{Dataset Analysis}

\subsubsection{Object categories}
\label{sec-obj}
\wzl{As we aimed to design VinT-6D, we established three primary selection criteria for the objects that would be used in the project. They should be easy to hold, commonly found in daily life, and made of various materials such as plastics, metal, and glass. This selection reflects the growing emphasis on domestic robotics applications, especially in healthcare and elderly assistance. At first, we selected 21 objects from the YCB-Video dataset \cite{calli2017yale}, considering their hand-held size and influence in robotics. However, we also recognized the limited availability and variety of objects, particularly outside the US. We added five more everyday objects with transparent and reflective features to address this, which you can find in \cref{obj-sim}. Then, we carefully selected 10 common items from them, such as a can of tomato soup or a bottle of mustard (\cref{obj-real}). Our main objective was to identify their functional requirements and the challenges that arise when handling them. These items include a metal shaker and a wine glass, which come in different sizes and materials, and some of them require a certain level of skill when it comes to rotating them or holding them with just the fingertips.}

\begin{figure*}[t!]
\vskip 0.2in
\begin{center}
\centerline{\includegraphics[width=\linewidth]{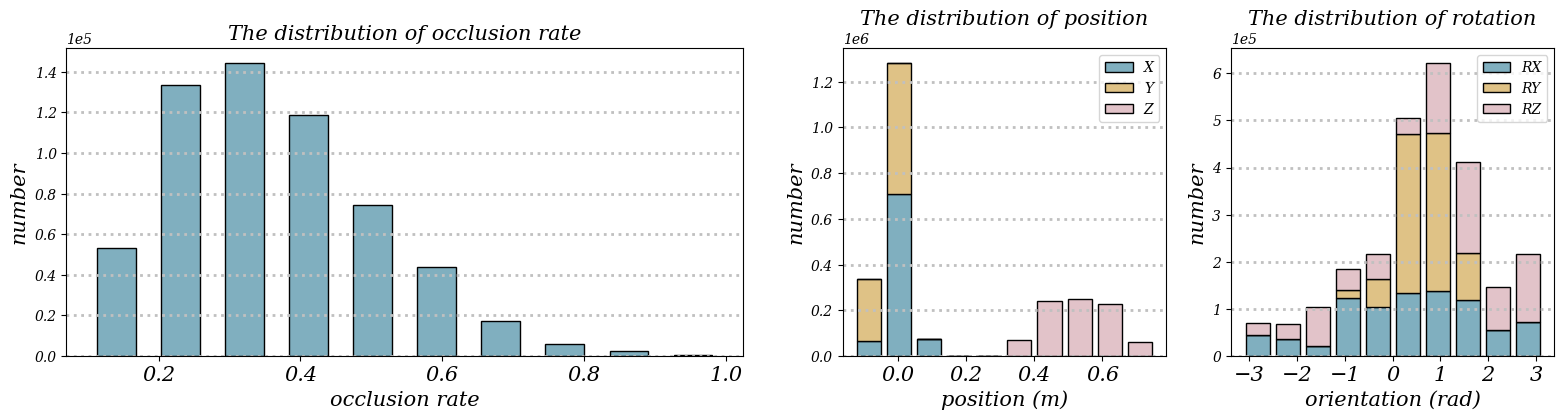}}
\caption{\textbf{Distribution of occlusion rate and camera poses in VinT-Sim.}}
\vspace{-3mm}
\label{statistics-sim}
\vspace{-7mm}
\end{center}
\end{figure*}

\begin{figure*}[t!]
\vskip 0.2in
\begin{center}
\centerline{\includegraphics[width=\linewidth]{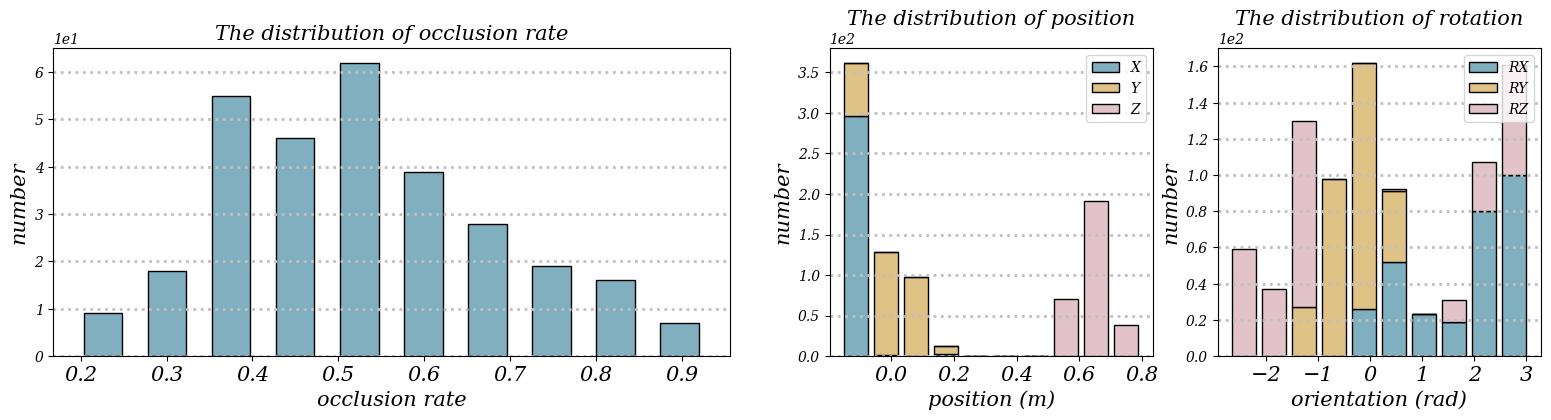}}
\vspace{-3mm}
\caption{\textbf{Distribution of and occlusion rate and object poses in VinT-Real.}}
\label{statistics-real}
\vspace{-3mm}
\end{center}
\vskip -0.2in
\end{figure*}

\subsubsection{\wzl{Generalization of Robotic Hands}}
\label{sec-hand}
\wzl{ The use of different multi-fingered hands in robotics presents unique challenges that must be addressed. Robotic vision, touch, and proprioception can vary significantly depending on the configuration of the hands used, as each configuration provides different proprioception information and activated taxels. Moreover, the size and appearance of the hand can also impact the robotic vision. To overcome these challenges and enhance the generalization of robotic hands, VinT-Sim incorporates two widely recognized types of hands - three-fingered and four-fingered. The Trx hand, with its three fingers, eight degrees of freedom, is our own custom-made solution. The Allegrohand, on the other hand, is a commercially available four-fingered hand with 16 degrees of freedom. In VinT-Real, we use the Trx hand at the end of the robot arm.}


\subsubsection{\wzl{Whole-Hand Tactile Perception}}
\label{sec-tactile}
\wzl{VinT-6D is a unique dataset that uses whole-hand tactile perception, which sets it apart from other datasets that rely on fingertip tactile sensing alone. By strategically placing array-based tactile sensors across the fingertips, pulp, and palm, this dataset offers comprehensive local information during object-hand interaction, particularly in scenarios where vision is obstructed. The Trx hand features 620 taxels, while the Allegro hand boasts 679 taxels, covering the entire hand. This promises to significantly advance robotic in-hand perception capabilities through extensive area contact.}



\subsubsection{\wzl{Diversity of Available Input Modalities}}
\wzl{VinT-6D provides researchers with various input modalities to aid further exploration. In VinT-Sim, researchers can access a range of visual inputs such as color images, depth images, and segmentation labels. Moreover, VinT-Sim also offers tactile point cloud and proprioception data, such as finger poses relative to the palm. In Vint-Real, we collect the left and right eye images from a stereo camera, depth and IR images from a Kinect Azure camera, touch readings from tactile sensors, finger joint angles from Trx-hand, and object and hand poses from the Vicon motion capture system, as demonstrated in \cref{bartender-robot}. These diverse modalities offer promising directions for specific in-hand perception like transparent object pose estimation and reconstruction or perceiving liquid using IR images.}


\subsubsection{\wzl{Well-Calibrated Robotic Platform}}
\wzl{Achieving high-quality data collection demands a well-calibrated robotic platform with all sensors aligned in timestamps and coordinate frames. Precise alignment of multi-sensors ensures accurate data collection, particularly when acquiring segmentation labels as explained in \cref{app-real-vision}. To guide the SAM modal \cite{kirillov2023segment} in object segmentation from hand, we assign `add' and `remove' prompts, which use touch and proprioception information to guide visual segmentation. Aligning the timestamps and coordinates of multi-sensors is crucial as it makes it possible to segment the object from the hand.} 

\begin{figure*}[t!]
\vskip 0.2in
\begin{center}
\centerline{\includegraphics[width=\linewidth]{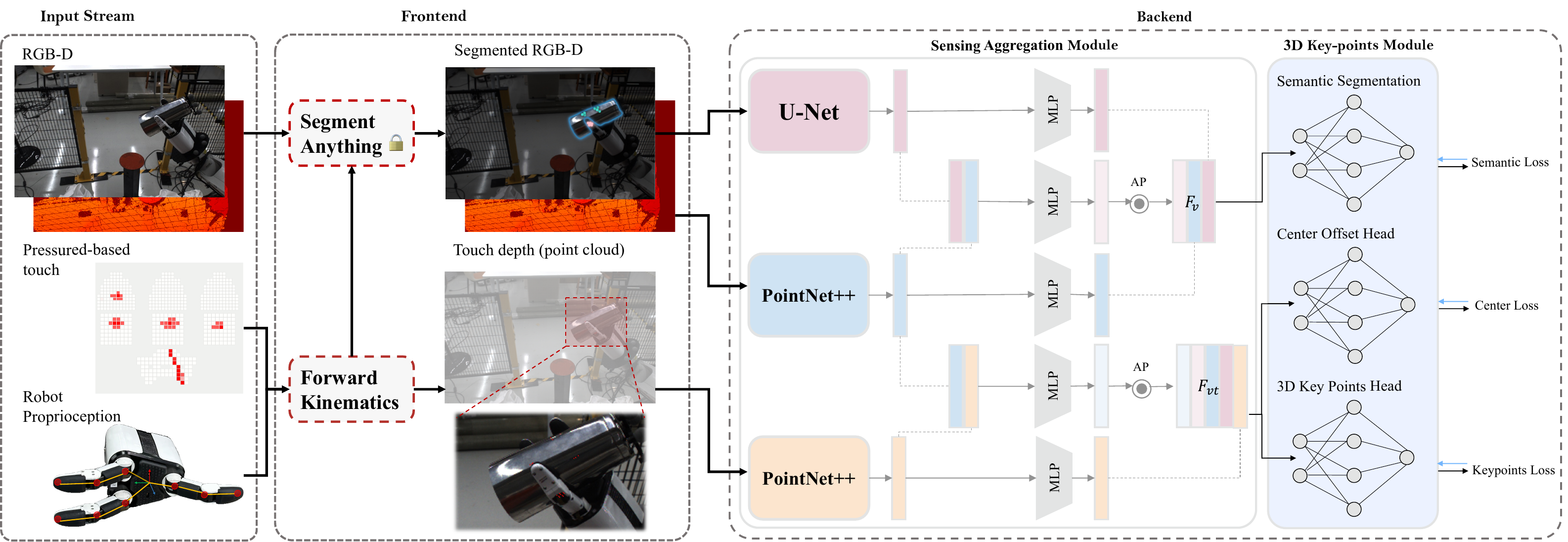}}
\vskip -0.1in
\caption{\textbf{Overview of VinT-Net.} The Frontend receives well-aligned vision and raw touch and proprioception data. Using SAM \cite{kirillov2023segment} and hand's forward kinematics, it acquires the segmentation label and touch point cloud separately. Then, the Backend predicts the object's pose. }
\label{vint-net-framework}
\end{center}
\vskip -0.4in
\end{figure*}

\subsubsection{\wzl{Robust Segmentation Towards Occlusion}}
\label{sec-stat-seg} 
Segmenting the in-hand object from vision remains a significant challenge.  
In our dataset, we concentrate on achieving robust object segmentation reliable object segmentation even when the robotic hand occludes a portion of the object. To address this, we have employed multi-modal prompts in SAM to differentiate between objects and hands. Specifically, we use negative (``-'') for thumb and positive (``+'') for index and middle finger tactile points, along with two object points. This mirrors typical object-in-hand poses collected in real-world scenarios. To evaluate SAM's reliability when dealing with occlusions ranging from 20\% to 50\% on a `tomato soup can' from VinT-Real, we conducted a comparison against manually segmented ground truth across 100 sets of data. The results are presented in \cref{Tab: Occlusion_SAM}.

\begin{table}[t!]
\caption{\textbf{Object-in-hand segmentation accuracy with different occlusion levels in VinT-Real.}}
\label{Tab: Occlusion_SAM}
\begin{center}
\begin{small}
\begin{sc}
\begin{tabular}{lcccr}
\toprule
occlusion rate & 20\%& 30\%& 40\%& 50\% \\
\midrule
miou & 96.41 & 94.44 & 96.63 & 95.65\\
\bottomrule
\end{tabular}

\end{sc}
\end{small}
\end{center}
\vskip -0.2in
\end{table}

\subsubsection{\wzl{Data Diversity}}
\label{sec-stat-sim}

\wzl{We created 2,500 interactions for each object in VinT-Sim, resulting in a wide range of poses. These poses were rendered in Blender across four different environments, with three camera positions for each. This led to a total of 2 million visual-tactile data sets. The dataset itself includes finger-induced occlusions ranging from 10\% to 90\%, as well as various camera poses (\cref{statistics-sim}) and object poses (\cref{statistics-real}).}

\wzl{To diversify VinT-Real, we densely sample within the robot's workspace, akin to a toddler's exploratory play. Our data collection mirrors this process, capturing object-in-hand scenarios using vision, touch, and proprioception. We establish 125 key positions relevant to the robot and camera workspaces in the Cartesian space and introduce 8 hand rotations at each, totaling 1000 unique poses. These poses, designed to avoid obstruction by fixtures with markers and to be solvable via inverse kinematics, are repeatedly adjusted across ten variations per object. This ensures a diverse touch point collection and poses variability. We control hand trajectories and arm movements through robotic motion planning, amassing 100,000 well-aligned data instances across 10 objects, encompassing vision, touch, and proprioception.}

\section{VinT-Net}
\wzl{We introduce VinT-Net to enhance research in robotic object-in-hand pose estimation and validate the proposed dataset. This network acts as a simple yet effective baseline for multi-finger object-in-hand pose estimation challenges. VinT-Net can efficiently process various inputs, such as RGB and depth images and local touch points derived from robotic tactile and proprioceptive sensors. The architecture of VinT-Net comprises two crucial sub-modules: a sensing aggregation module (refer to \cref{fusion}) and a 3D keypoint-based pose estimation module (see \cref{estimator}). The framework of VinT-Net is illustrated in \cref{vint-net-framework}. It accurately estimates the object's 6D pose from the robot's camera perspective.}

\subsection{Sensing Aggregation Module}
\label{fusion}
\wzl{In the Sensing Aggregation Module, our focus is on leveraging color images, depth images, and locally computed touch data derived from proprioceptive and touch sensors. Utilizing the robot's forward kinematics, detailed in \cref{vint-net-framework}, we obtain the precise local touch point data $D_{touch}$. The module employs three separate branches for processing: the color vision data $D_{color}$ undergoes processing via a U-Net-based model \cite{ronneberger2015u} for appearance feature extraction. Simultaneously, the depth vision data $D_{depth}$ and the local touch data $D_{touch}$ are processed using individual PointNet++ \cite{qi2017pointnet++} networks, isolating local and global geometric features. Following the principles of advanced vision fusion techniques \cite{wang2019densefusion}, we integrate the color and depth information on a pixel-wise level. The local touch data, treated as a complementary element to the depth image, is fused with the depth features to produce a comprehensive visual feature $F_{v}$ and a fused visual-tactile feature $F_{vt}$, as detailed in \cref{vint-net-framework}.}

\subsection{3D Keypoint-Based Pose Estimation Module}
\label{estimator}

\wzl{It has been observed that in pose estimation of rigid objects, the 3D spatial relationship between any two points on the object remains constant, regardless of the object's movements. This observation has been noted in \cite{he2020pvn3d, he2021ffb6d}. As a result, it has been found that predicting selected 3D key points on the object based on visible surface appearances in a camera view is effective. However, occlusions can significantly impair vision when a robotic hand grasps an object. To address this issue, our method uses fused multi-modal embeddings to predict both the object's central point and its keypoint offsets. A multi-modal fusion module is used to extract per-point features from vision and touch inputs, which are then fed into three MLP heads. These heads are responsible for predicting semantic labels, center, and keypoint offsets. Interestingly, our experiments showed that integrating touch data into the semantic segmentation head did not enhance performance. This may indicate that additional touch information may interfere with the head's functionality. We apply Focal Loss \cite{lin2017focal} for semantic segmentation supervision and $L1$ Loss for center and 3D keypoint offset prediction. A combined multi-task loss augments the overall model accuracy :}
\begin{equation}
    L_\text{multi-task }{ = }\lambda_1L_\text{semantic}+\lambda_2L_\text{ keypoints}+\lambda_3L_\text{center}
\label{equa-1}
\end{equation}
\wzl{where $\lambda_1$, $\lambda_2$ and $\lambda_3$ are the weights for each task.}

\wzl{In the final step, following \cite{he2020pvn3d, he2021ffb6d}, a clustering algorithm is employed to distinguish between instances with identical semantic labels. A least-squares fitting algorithm accurately determines the 6D pose parameters.}
\vspace{-3mm}
\begin{equation}
    L_\text{least-squares}=\sum_{j=1}^M||kp_j-(R\cdot kp_j^{^{\prime}}+t)||^2
\end{equation}
where $M$ represents the pre-selected 3D keypoints of an object, $R$ symbolizes rotation, and $t$ represents translation.

\section{Experimental Results}

\label{sec-exp}

\subsection{Experimental Setup}

\textbf{Implementation Details.} \wzl{We used the Adam optimizer with an initial learning rate of 0.01 for training and set the batch size at 24. The training process was conducted over 25 epochs, and we set the hyper-parameters $\lambda_1$, $\lambda_2$, and $\lambda_3$ to 1, 2, and 1, respectively. The training and testing processes were executed on a computing server equipped with 6 Quadro RTX 8000 GPUs. The VinT-Sim synthesis procedures were conducted by a cloud computing platform that utilized 16 NVIDIA P40 GPUs.}

\textbf{Evaluation Metrics.} \wzl{In evaluation, we follow \cite{he2020pvn3d} and evaluate our method with the ADD and ADD-S metrics, as defined in \cite{xiang2017posecnn}. The ADD metric measures the average distance between the object's vertices transformed by the predicted 6D pose $\left[R,t\right]$ and the corresponding vertices transformed by the ground truth pose $[R^*,t^*]$.} 
\begin{equation}
    \text{ADD}=\frac1m\sum_{x\in\mathcal{O}}||(Rx+t)-(R^*x+t^*)||
\end{equation}
where $x$ is one of the $m$ vertexes on the object mesh $\mathcal{O}$. The ADD-S metric is used for symmetrical objects; it computes the average distance based on the closest point distance. 
\begin{equation}
    \text{ADD-S}=\frac1m\sum_{x_1\in\mathcal{O}}\min_{x_2\in\mathcal{O}}||(Rx_1+t)-(R^*x_2+t^*)||
\end{equation}
We compute the ADD(S) AUC, the area under the accuracy-threshold curve, which is obtained by varying the ADD distance threshold for non-symmetric objects and ADD-S distance for symmetric objects.

\vspace{-3mm}
\subsection{Evaluation on VinT-6D}

\begin{table}[t!]

\caption{\textbf{Quantitative Evaluation on VinT-6D Dataset.} ADD(S) AUC metric is reported.}
\label{Tab: performance on vint-6d}

\begin{center}
\begin{small}
\begin{sc}
\begin{tabular}{lcccr}
\toprule
 &sim&real&vint-6d&\\
\midrule
blue bottle & 87.52& 93.45& 94.15\\
large shaker & 88.04& 94.66& 96.23\\
stick & 78.04& 83.72& 83.76\\
potted meat can & 88.56& 94.31& 95.53\\
tomato soup can & 89.77& 94.65& 95.44\\
power drill & 88.04& 91.72& 97.09\\
tuna fish can & 90.54& 92.34& 93.89\\

\bottomrule
\end{tabular}

\end{sc}
\end{small}
\end{center}
\vskip -0.2in
\end{table}
\wzl{We conduct experiments to compare the performance of models trained on purely simulated data, real-world data, and a combination of both. The results of these experiments are presented in \cref{Tab: performance on vint-6d}. The results show that our comprehensive calibration and simulation strategy effectively bridges the sim2real gap by augmenting the simulated data to the real data, as evidenced by the performance improvement when combining simulated and real data for training.}

\vspace{-3mm}
\subsection{A Comprehensive Analysis}
\label{Ablation}
We first ablate the benefits of touch and proprioception on the dataset, then compare the results with other object-in-hand pose estimation methods, and finally evaluate the robustness under various occlusion levels.
\vspace{-4mm}
\paragraph{Benifit of touch and proprioception.} \wzl{This ablation study evaluates the contributions of tactile and proprioception modalities in VinT-Real. In \cref{ablation-table}, we compare our visual-tactile Vint-Net with the visual baseline \cite{he2020pvn3d}. Our visual-tactile VinT-Net surpasses the baseline relying solely on vision by a margin, demonstrating the incorporation of additional tactile information notably enhances performance. 
} 
\begin{table}[t!]

\vskip 0.15in
\caption{\textbf{Ablation Studies of Vision and Touch in VinT-Real.} `Touch' comprises touch and proprioception data.}
\label{ablation-table}
\begin{center}
\begin{small}
\begin{sc}
\begin{tabular}{lcccr}
\toprule
objects & vision & vision + touch \\
\midrule
blue bottle & 90.34 & 93.45\\
large shaker & 87.04 & 94.66\\
stick & 78.04& 83.72\\
potted meat can & 91.23&94.31 \\
tomato soup can & 86.78& 94.65\\
power drill & 90.04 & 91.72\\
tuna fish can & 85.81& 92.34\\
\bottomrule
\end{tabular}
\end{sc}
\end{small}
\end{center}
\vspace{-0.3in}
\end{table}

\vspace{-4mm}
\paragraph{Comparison to other object-in-hand pose estimation methods.} \cref{Tab: Comaprsion_sota} shows that our VinT-Net significantly outperforms recent object-in-hand pose estimation methods like \cite{wen2020robust} on the ADD-0.05d metric, which notably employs hand state estimation to enhance pose accuracy. To ensure a fair comparison given hardware differences, we reproduced the Object-Hand-Pose method by replacing its two-finger gripper with our three-finger hand, which has reduced degrees of freedom. This and other adjustments were made to estimate the `tomato soup can' pose within our VinT-Real dataset, as shown in \cref{Tab: Comaprsion_sota}.

\begin{table}[t!]

\caption{\textbf{Comparison to other object-in-hand methods on `tomato soup can'.} ADD-0.05d metric is reported.}
\label{Tab: Comaprsion_sota}
\begin{center}
\begin{small}
\begin{sc}
\begin{tabular}{lcccr}
\toprule
methods & add& \\
\midrule
pvn3d \cite{he2020pvn3d} & 74.60& \\
object-hand-pose \cite{wen2020robust} & 76.74& \\
ours & 82.43& \\

\bottomrule
\end{tabular}
\end{sc}
\end{small}
\end{center}
\vskip -0.2in
\end{table}

\vspace{-4mm}
\begin{table}[t!]

\caption{\textbf{Robustness to occlusion by the multi-fingered hand on `tomato soup can.'} ADD-0.05d metric is reported.}
\label{Tab: Comaprsion_occlusion}
\begin{center}
\begin{small}
\begin{sc}
\begin{tabular}{lcccr}
\toprule
occlusion rate& 20\% & 30\% & 40\% & 50\% \\
\midrule
vision only& 93.31& 89.45 & 85.77 & 80.43\\
vision + touch& 94.76& 93.89 & 90.47 & 88.81\\

\bottomrule
\end{tabular}
\end{sc}
\end{small}
\end{center}
\vskip -0.2in
\end{table}

\paragraph{Robustness to occlusion by the multi-fingered hand.} To prove the robustness against occlusion caused by the multi-fingered hand, we further investigate the `Tomato Soup Can' under different occlusion levels. \cref{Tab: Comaprsion_occlusion} reveals a notable decrease in accuracy for vision-based approaches as occlusion intensifies, from 93.31\% at 20\% occlusion to 80.43\% at 50\% occlusion. Conversely, the multi-modal fusion strategy, which leverages both vision and tactile inputs, exhibited remarkable resilience against increasing occlusion levels. Its accuracy slightly dipped from 94.76\% at 20\% occlusion to 88.81\% at 50\% occlusion.




\vspace{-2mm}
\section{Conclusion}
\vspace{-2mm}
We contribute VinT-6D, a pioneering multi-modal dataset for 6D object-in-hand pose estimation, by integrating vision, touch, and proprioception. With over 2 million and 0.1 million synthesized (VinT-Sim) and real data (VinT-Real), VinT-6D is tailored for robotic hands, providing high-quality, well-aligned data for accurate object-in-hand pose estimation. Our benchmark method leveraging VinT-6D showcases notable performance improvements, highlighting the dataset's potential to bridge the gap between simulated and real-world applications. 

\section{Limitations and Future Work}
While we have endeavored to narrow the sim2real gap, we acknowledge that our existing VinT-6D dataset lacks diversity in terms of objects and scenes. As part of our continuous research efforts, we intend to introduce a broader range of elements and more varied object-in-hand grasping scenarios in the future. Furthermore, our proposed VinT-Net represents an instance-level pose estimation approach that employs a simple fusion strategy for dataset validation. This leaves a large space to explore category-level or zero-shot object-in-hand pose estimation, which could potentially benefit from a more advanced fusion of multiple modalities.

\section*{Impact Statement} 
The imminent release of VinT-6D marks a substantial contribution to the field, promising to enhance research and development in robotic manipulation.

\section*{Acknowledgement}

This research was supported by Tencent and partially by the National Natural Science Foundation of China under Grant 62173352. We would like to express our gratitude to all the staff at Robotics X Lab who contributed to this dataset, and to Zida Zhou for his fixture design.

\nocite{langley00}

\bibliography{example_paper}
\bibliographystyle{icml2024}

\newpage
\appendix
\onecolumn
\section{Appendix.}
\subsection{VinT-Sim}


\subsubsection{Simulation setup}
Given the influence of finger configurations on visual and tactile data acquisition, we have included datasets for both three-fingered and four-fingered hand configurations in our simulation. We use our three-fingered TRX-Hand and Wonik's four-fingered AllegroHand, both equipped with pressure-based touch sensors on each finger.

\subsubsection{Simulating Touch}
\label{app-touch-sim}
To narrow the sim-to-real gap in touch simulation, we pose two critical questions: (1) How can we represent tactile readings in a sensor-agnostic manner, accounting for different types of tactile sensors and their performance variation over time? (2) How can we accurately simulate the distribution of all taxels (tactile unit sensors)?

To tackle these queries, we first reviewed recent pioneering approaches in tactile simulation, primarily using: (1) \cite{suresh2023neural, qi2023general, smith20203d, xu2023visual} Directly using the simulator \cite{wang2022tacto} to generate RGB images for vision-based tactile sensors. (2) \cite{dikhale2022visuotactile, li2023vihope, rezazadeh2023hierarchical} pressure-based tactile sensors simulated by attaching depth cameras to each joint of an AllegroHand, with the point cloud serving as tactile feedback. However, these approaches may introduce a significant domain gap compared to real-world tactile sensors, primarily because they do not account for the curved surface of fingers and the typically smaller, more localized actual touch area. 

\begin{figure}[t!]
\vskip 0.2in
\begin{center}
\centerline{\includegraphics[width=\columnwidth]{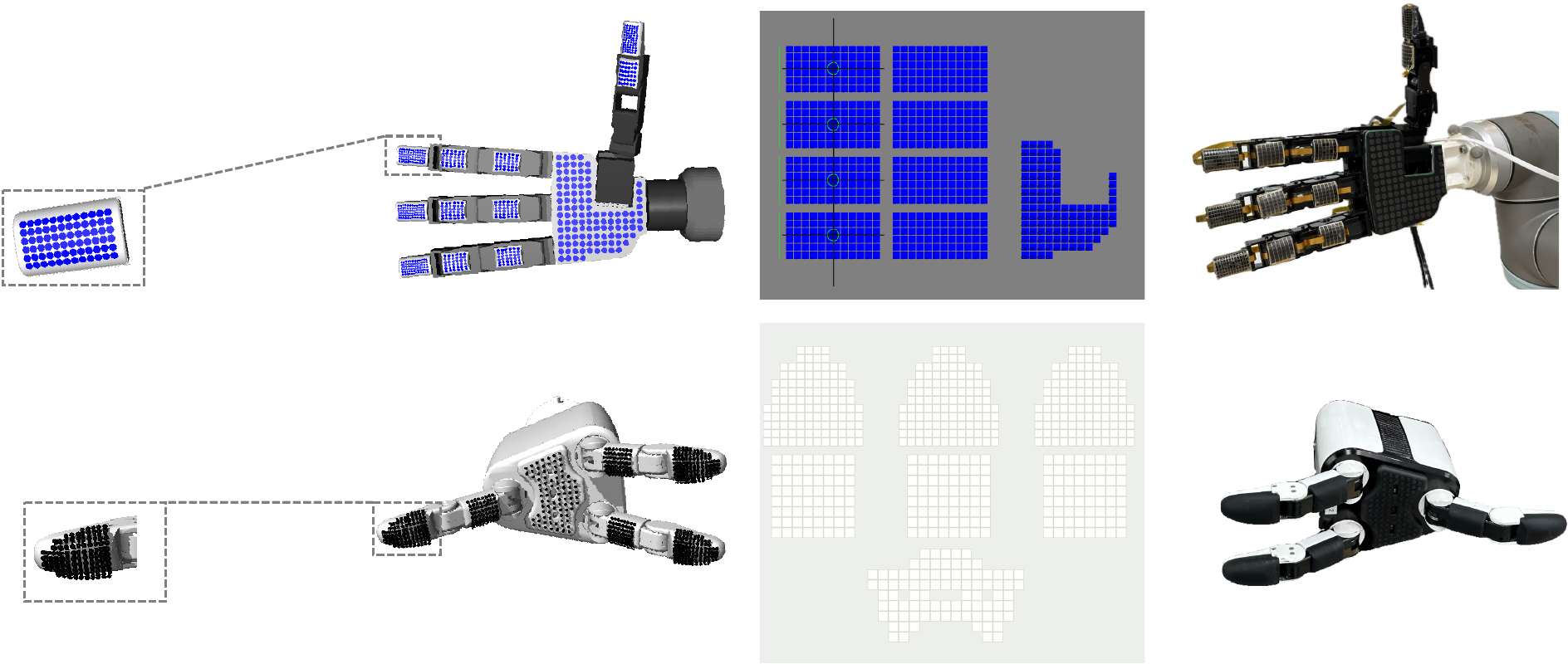}}
\caption{\textbf{Simulating touch follows the real-world setup.} In the upper row, we have an AllegroHand equipped with whole-hand tactile sensors, while in the lower row, we have a Trx-hand with sensors. The distribution of these sensors closely follows that of the real-world setup.}
\label{touch-sim-1}
\end{center}
\vskip -0.2in
\end{figure}

In our real-world setup, as shown in \cref{touch-sim-1}, each finger and the palm of the robotic hands are equipped with tactile sensors featuring a piezoresistive array of taxels on a curved, continuous surface. The three-fingered hand has 620 taxels, while the four-fingered hand comprises 679 taxels. Unlike traditional flat tactile arrays, these curved surface sensors distribute tactile elements over an irregular surface, and slight deviations exist in each sensor's tactile element distribution due to manufacturing precision constraints. The response of each taxel is proportional to the applied normal force, displaying a non-linear, positively correlated response pattern. Over time, a greater force is required for the same level of activation.

For individual taxels, we meticulously modeled each taxel as a force sensor within the 3D robotics MuJoCo physics engine \cite{todorov2012mujoco}. Each simulated tactile element is represented as a small, lightweight cuboid cylinder on the finger surfaces, topped with a blue hemisphere, as depicted in \cref{touch-sim-1}. Upon tactile contact, the simulation generates a touch signal ($T$) and corresponding sensor readings ($R$) at the contact point. These hemispheres, when activated, denote the local surface topology being touched, facilitating interaction with the tactile sensors. Using forward kinematics, the activated location is computed as the touch signal, providing a versatile representation suitable for both pressure-based and vision-based sensors, and accommodating time-varying characteristics.

We aimed to replicate these tactile distributions accurately in our simulations for the overall distribution of taxels. We captured detailed mesh and texture information by meticulously scanning each sensor part with 3D scanning equipment. After exporting the calibrated positional and normal data for all tactile elements, we precisely aligned each element on the hand in the simulation. This approach mirrors real sensor features and distributions, effectively bridging the gap between simulated and real-world environments. We recorded all activated positions as a local touch point cloud ($D_{touch}$) relative to the palm frame, included for use in the dataset.

\subsubsection{Simulating Object-Grasp Interaction}
\label{app-obj-grasp}
What types of interactions are essential between a robotic hand and an object it touches? How to narrow down the sim2real gap in proprioception?
Our approach to selecting object-grasp interactions is guided by two primary considerations: the ability to hold the object stably in hand and the readiness of the grasp pose for subsequent robotic manipulation tasks. We posit that an effective in-hand grasp should prevent the object from dropping and position it optimally for future manipulative actions.

Notably, recent advancements in state-of-the-art multi-finger object-grasp generation, such as those proposed by \cite{turpin2023fast} and \cite{wan2023unidexgrasp++}, have emerged. 
However, these approaches often do not address the critical aspect of stable holding in hand and readiness for manipulative actions, which are key focuses of our research.

\begin{figure}[t!]
\vskip 0.2in
\begin{center}
\centerline{\includegraphics[width=\columnwidth]{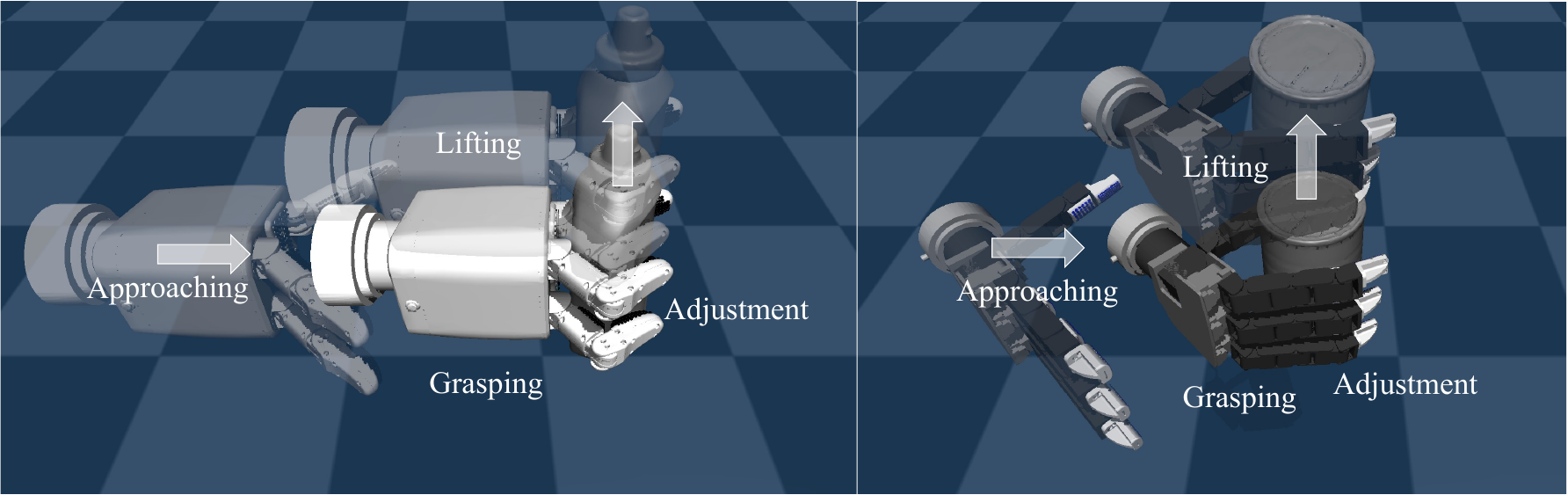}}
\caption{\textbf{Simulating object-grasp interaction}. Selected stable grasping requires stability and preparedness for manipulation. Our simulation-based grasping method replicates real-world scenarios to ensure effective tactile engagement.}
\label{mujoco-sim}
\end{center}
\vskip -0.5in
\end{figure}

\textbf{Simulating object-hand setup:} To simulate object-grasp interactions that adhere to these specific criteria, we have implemented a setup involving a robotic hand and an object within the MuJoCo physics engine \cite{todorov2012mujoco}. The detailed grasping process is depicted in \cref{mujoco-sim}. Initially, the Trx-Hand (or AllegroHand) and the object are placed into the simulation environment in predefined initial poses. The robotic hand is carefully positioned into a pre-grasp stance, ensuring its palm is tangentially aligned with a randomly chosen point adjacent to the object’s graspable surface. The proximity between the hand in its pre-grasp position and the object is assigned randomly, adhering to a set ratio. Utilizing inverse kinematics, we articulate the finger joints to facilitate contact between the tactile sensors on the fingers or palm and the object's surface. 
Upon establishing contact, 
the robotic hand executes a power grasp and lifts the object to a predetermined height.
This action is dependent on the successful and physically realistic integration of the object being held.
The grasping process is repeated for each combination of object and hand until a dataset comprising 2,500 successful grasps, each marked by effective tactile contact, is compiled. 
This setup is designed to closely mirror real-world grasping scenarios, ensuring the grasp's stability and utility for subsequent manipulations.
We meticulously record the object pose ($P_{obj}$), hand pose ($P_{hand}$), and the poses of all finger knuckles ($P_{knuckle}$) for subsequent vision rendering.

\subsubsection{Simulating Vision}
\label{app-sim-vision}
Drawing from our prior experience, we have recognized that images captured in MuJoCo for training frequently exhibit a significant domain gap. This issue predominantly arises from naive renderings, particularly for transparent or reflective objects. 
To obtain more realistic visual representations
of object-hand interactions, we utilize Blender to render two types of images: an RGB image and
a depth image, both depicting an object occluded by the
robotic fingers grasping it.

\begin{figure*}[t!]
\vskip 0.2in
\begin{center}
\centerline{\includegraphics[width=\linewidth]{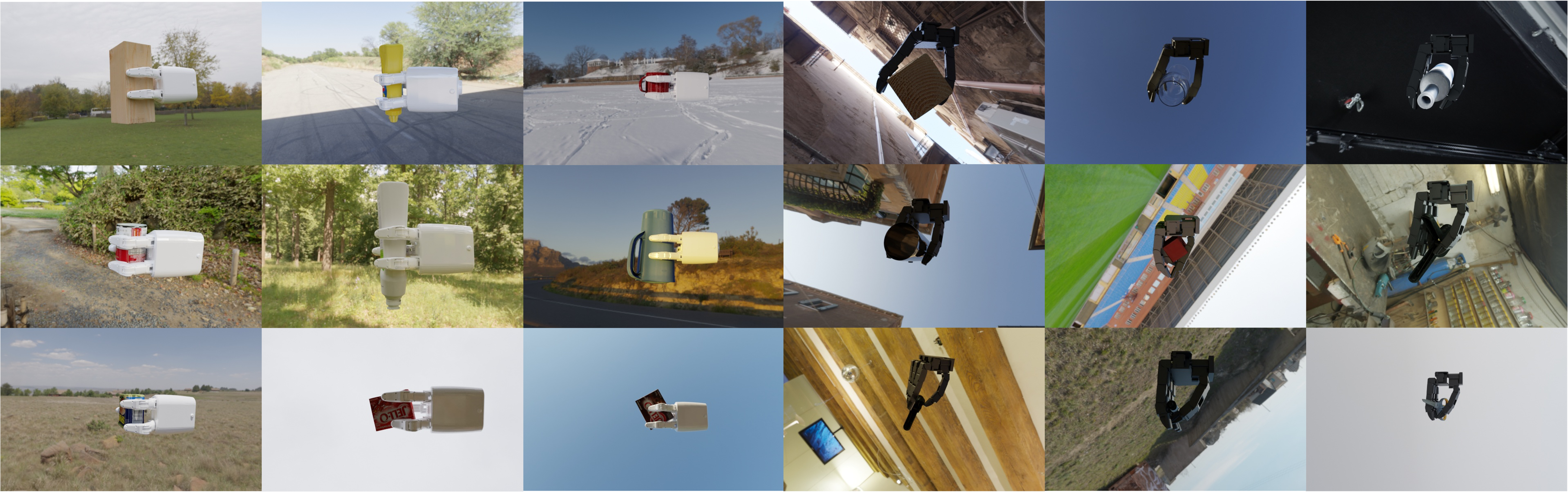}}
\caption{\textbf{Photorealistic Rendered RGB Images.} We present a set of photo-realistic rendered images that depict objects held in hand.}
\label{app-render-in-sim}
\end{center}
\vskip -0.2in
\end{figure*}

\textbf{Simulating RGB images}.
We employ Blender's integrated Cycles rendering engine to produce photorealistic scenes, each featuring an object held in hand. Initially, we import both an object and a hand model into Blender, setting their poses as well as the poses of all finger knuckles based on the information recorded during the object-grasp interaction phase.  For each object-in-hand configuration, we opt for different HDRI backgrounds \cite{polyhaven} to ensure adequate and realistic illumination and diverse backdrops. We render each scene from multiple viewpoints within a hemispherical range, focusing on the object-in-hand configuration. To achieve realistic rendering, as shown in \cref{app-render-in-sim}, we restrict the number of ray bounces and the number of glossy reflection bounces.

\begin{figure}[t!]
\vskip 0.2in
\begin{center}
\centerline{\includegraphics[width=\columnwidth]{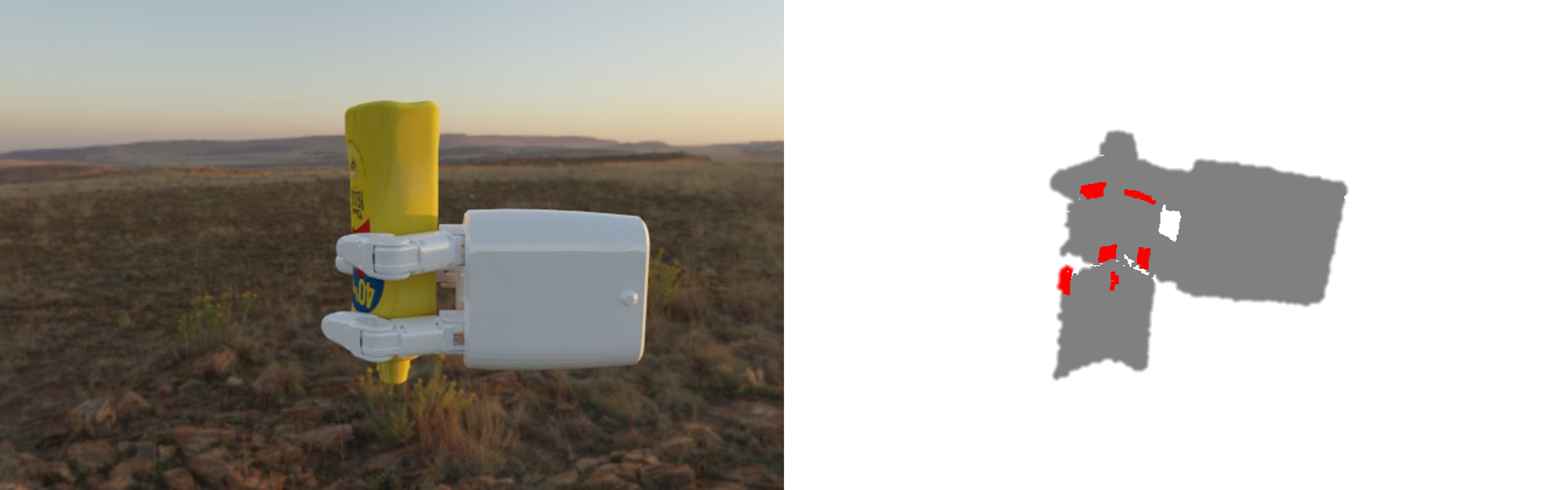}}
\caption{\wzl{\textbf{Well-Aligned Vision and Touch Data Visualization.} The image on the left shows a color-rendered image of the object being held, while the right side depicts the point cloud generated by the depth camera. The gray points represent the depth, while the red points indicate the touch points that have been activated.}}
\label{vision-touch-align-sim}
\end{center}
\vskip -0.2in
\end{figure}

\textbf{Simulating Depth images}. From a given camera viewpoint, a depth image was initially rendered using Blender's internal depth estimation.
In our real-world setup, we equipped the real-world robot with a Kinect Azure TOF camera. The depth readings from this camera are subject to various factors, including the distance between the camera and the object, environmental lighting conditions, and the object's surface specularity. Even under optimal conditions, the depth measurements from the Kinect Azure are influenced by a multitude of hyperparameters, significantly impacting the accuracy of the depth ground truth. To replicate these real-world conditions in our simulation, we introduced noise, created holes, and applied smoothing to the rendered depth image. This methodology, following the approach outlined by \citet{tolgyessy2021evaluation}, is specifically adapted to emulate the unique noise characteristics of the Kinect Azure sensor. We convert the depth image for visualization into a point cloud, aligning it with the local touch point cloud at the camera frame, as depicted in \cref{vision-touch-align-sim}.

\subsection{VinT-Real}
\subsubsection{Acquiring Equipment}

\begin{figure*}[t!]
\vskip 0.2in
\begin{center}
\centerline{\includegraphics[width=\linewidth]{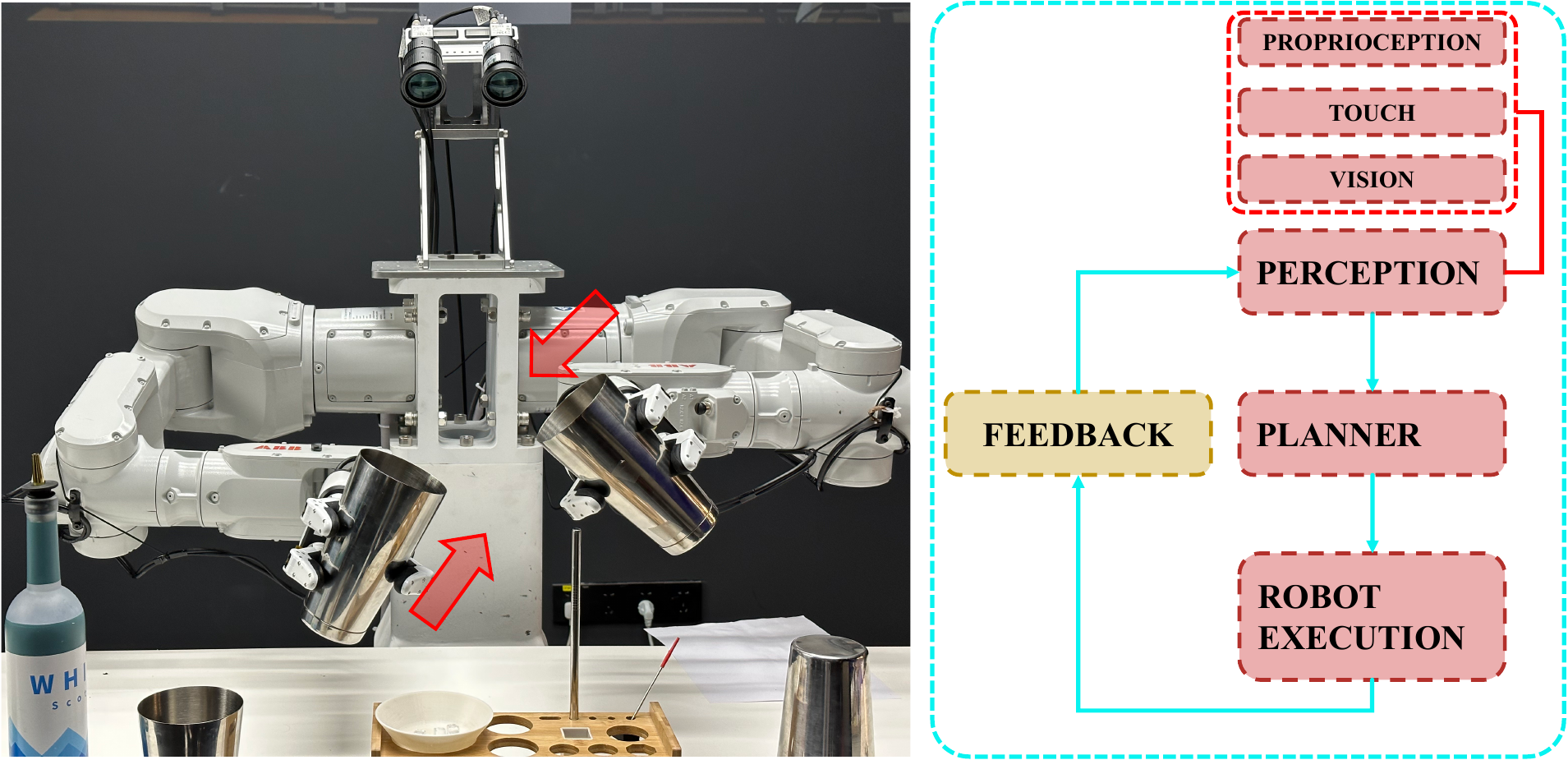}}
\caption{\textbf{Robotic Pouring Task.} Demonstrates the ``Perception-Planning-Control" paradigm in a robotic pouring task, highlighting challenges in unstructured household environments due to potential vision obstruction by the robot's hand. Essential for success is the integration of tactile feedback from finger and palm sensors and proprioceptive data from arm and hand joints.}
\label{app-bartender}
\end{center}
\vskip -0.2in
\end{figure*}

\begin{figure}[t!]
\vskip 0.2in
\begin{center}
\centerline{\includegraphics[width=\columnwidth]{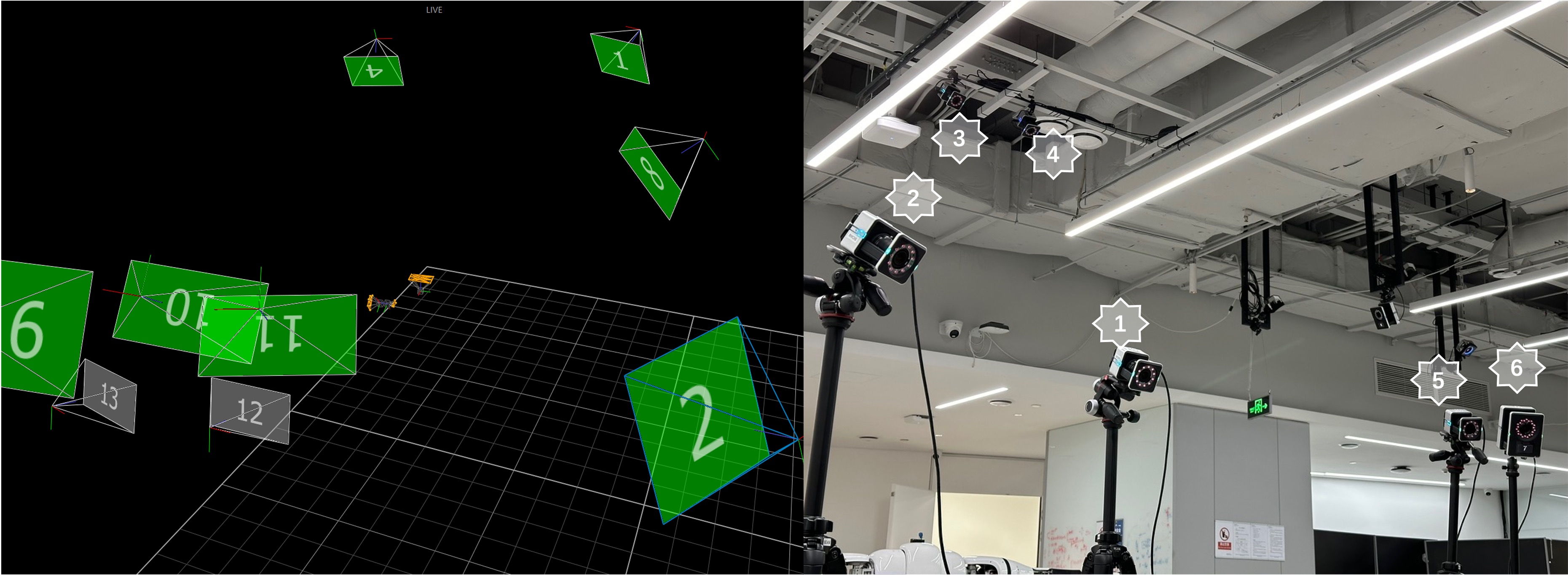}}
\caption{\textbf{Motion Capture System.} The motion capture system accurately captures the poses of the object, hand, and robot base.}
\label{app-vicon}
\end{center}
\vskip -0.2in
\end{figure}

As illustrated in \cref{app-bartender}, our customized robot platform integrates a comprehensive perception system, including vision, tactile, and proprioceptive systems. The vision system features two industrial color cameras for binocular stereo vision, enhanced by a TOF depth camera from Azure Kinect placed strategically between them. For tactile sensing, we have developed our own pressure-based sensors embedded in the fingers and palms. These self-developed sensors, uniquely designed for irregular surfaces, respond rapidly at frequencies exceeding 250 Hz, providing nuanced tactile feedback. The proprioceptive capabilities are enabled by two calibrated ABB arms and Trx-Hands, which precisely track movements of both arm and finger joints. Additionally, our configuration includes a motion capture system\footnote{https://www.vicon.com/software/tracker/} (\cref{app-vicon}), ensuring highly accurate tracking of objects and hand poses.

\subsubsection{Acquiring Touch}
\label{app_acquire_touch}
\textbf{Acquiring touch point cloud}.As highlighted earlier, the Trx-hands are outfitted with pressure-based sensors, as illustrated in \cref{touch-sim-1}. Each hand is equipped with sensors that yield 620 values from various points across the knuckles and palm, providing precise contact pressure readings at each touch point.
Our bespoke tactile sensors, designed to mimic human fingers' dexterity, can conform to irregular surfaces. Each sensing point on these sensors demonstrates remarkable responsiveness, which is evident through their low activation force, extensive measurement range, and high sampling frequency.
During the simulation phase, touch positions are aggregated to form a local touch point cloud. However, translating this into a real-world application presents significant challenges due to inherent system biases.

We initially calculate the hand pose relative to the robot's base frame using forward kinematics to address this. This calculation is extended from the finger knuckles to the hand frame. We then identify the activated touch sites and map them onto the finger knuckles' positions to form the local touch point cloud. Given the complex series of calculations involved in proprioception, achieving the necessary precision in the touch point cloud requires a highly calibrated robotic system. While one common method to minimize hand pose errors involves attaching ArUco tags \cite{garrido2014automatic}, as demonstrated in \cite{dikhale2022visuotactile}, \cite{li2023vihope}, \cite{rezazadeh2023hierarchical}, and \cite{xu2023visual}, this approach can lead to centimeter-level inaccuracies, which fall short of our precision requirements.

To overcome this, we equipped the Trx-hand with a flange plate featuring markers and utilized a motion capture system to record the hand pose, achieving sub-millimeter accuracy. Furthermore, both the hand and arm were rigorously calibrated. The complete touch points were visualized and meticulously aligned with the vision system, as detailed in \cref{vision-touch-align}.

\subsubsection{Acquiring Vision}
\label{app-real-vision}
\textbf{Aligning RGB-D images}. This process necessitates aligning the depth images with the RGB images. We obtain RGB images from the right camera of the stereo setup, which has a resolution of 1200 x 960, and depth images from the central TOF camera with a resolution of 640 x 565. Using a chessboard, we repeatedly recalibrated both cameras' intrinsic and extrinsic parameters to ensure precise alignment. After distortion correction, we reproject the depth image into the right camera's frame, resulting in two well-aligned vision images with resolutions of 640 x 400,.


\begin{figure}[t!]
\vskip 0.2in
\begin{center}
\centerline{\includegraphics[width=\columnwidth]{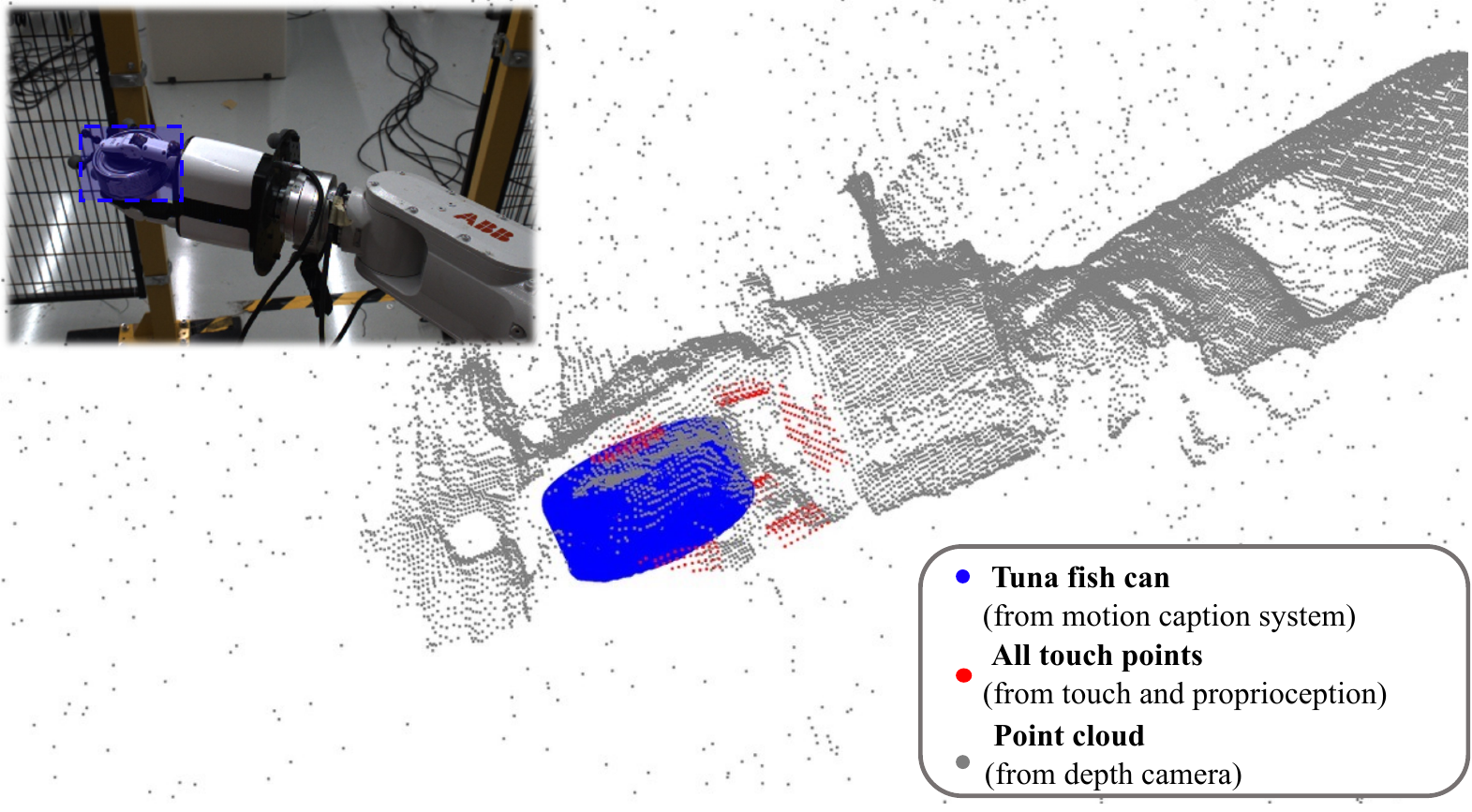}}
\caption{\textbf{Visualization of full tactile points aligned with vision in VinT-Real.} Gray points represent the point cloud from the depth camera, blue points represent the transformed model from the motion capture system, and red points indicate the full touch points of the hand.}
\label{app-vision-touch-align}
\end{center}
\vskip -0.2in
\end{figure}

\begin{figure}[t!]
\vskip 0.2in
\begin{center}
\centerline{\includegraphics[width=1\linewidth]{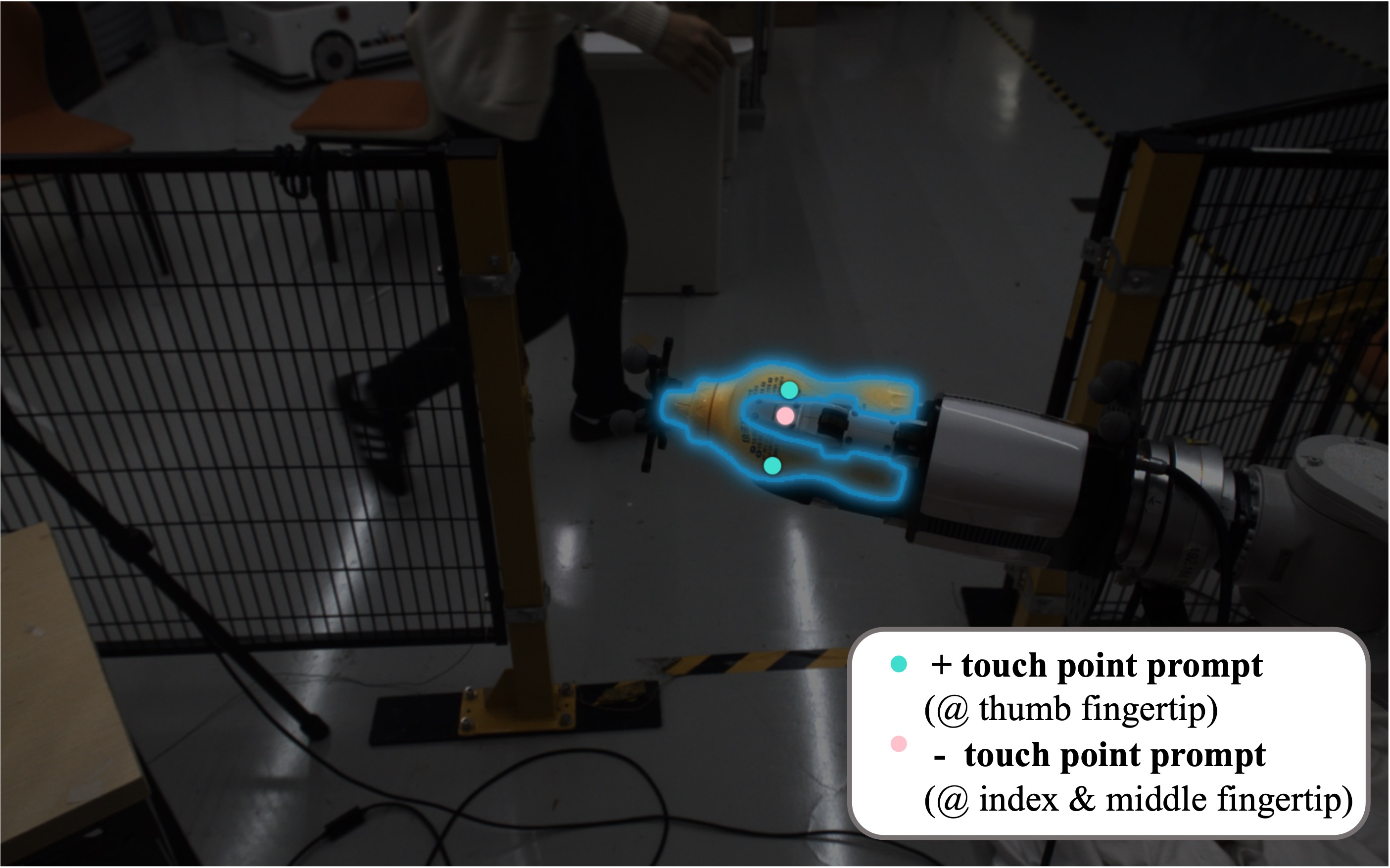}}
\caption{\textbf{Object-in-hand segmented by large vision model. We use touch and proprioception as cues for the segmentation of a large model. We assign `add' prompts to the activated touch points on the tip of the thumb fingertip while assigning `remove' prompts to those on the index and middle fingertips. This approach helps us to effectively and efficiently segment the model.}}
\label{seg-by-sam}
\end{center}
\vskip -0.2in
\end{figure}

\textbf{Acquiring segmentation labels}. 
When an object is held in hand, it is often obscured by occlusions caused by multiple fingers, tightly intertwining the object and the hand. This complexity poses a significant challenge in segmenting the object from the hand in aligned images. Despite the advancements in large-scale segmentation models, like \cite{qi2023high} and \cite{kirillov2023segment}, segmenting objects in hand still necessitates specific prompts for guidance. Manual hover-and-click methods for adding and removing prompts in our extensive dataset are impractical. Our pivotal insight involves using touch and proprioception as cues for SAM \cite{kirillov2023segment}. With our vision and touch data precisely aligned to millimeter accuracy, we leverage touch data and proprioception for segmentation. Specifically, using the activated touch points and the camera matrix, we calculate the pixel position of each touch point through robotic kinematics. As demonstrated in \cref{seg-by-sam}, we assign `add' prompts to activated touch points on the thumb fingertip and `remove' prompts to those on the index and middle fingertips. We employed the ViT-H SAM model with 636 million parameters for our experiments.

\subsubsection{Acquiring Object-in-hand Pose Gound Truth}
\label{app-real-pose}
\begin{figure}[t!]
\vskip 0.2in
\begin{center}
\centerline{\includegraphics[width=\columnwidth]{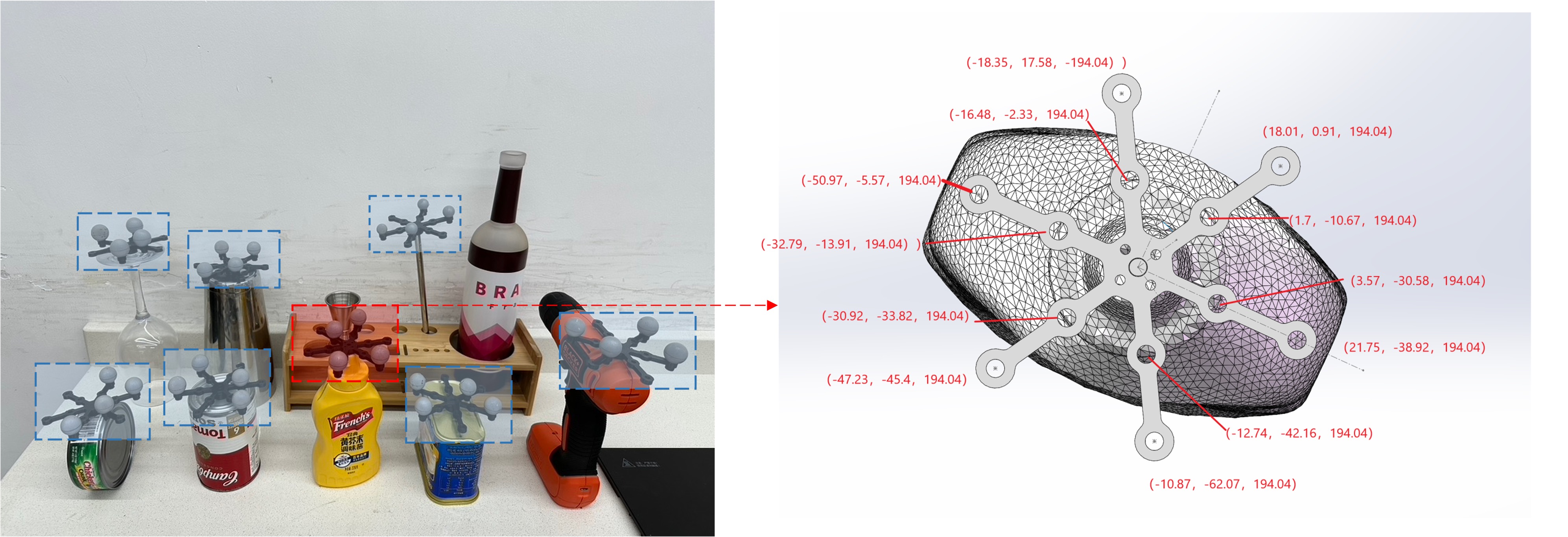}}
\caption{\textbf{Real-world objects with markers.}}
\label{object-markers}
\end{center}
\vskip -0.2in
\end{figure}

Obtaining ground truth for object-in-hand poses is straightforward in simulations but presents significant challenges in the real world. Unlike object pose estimation datasets in computer vision, such as those in \cite{calli2017yale} where objects are simply placed on a table and their poses estimated via QR codes, we adopt a motion capture system similar to \cite{dikhale2022visuotactile}, which attaches markers directly to the objects. However, we believe that simply sticking markers on objects may not yield high-quality pose data due to the uncertainty in marker positions relative to the object frame. To address this, we have custom-designed fixtures for each object, affixing markers on these fixtures. We then securely attach these marker-equipped fixtures to the objects, ensuring that the anchoring is both visually and physically feasible when the object is held in hand, as illustrated in \cref{object-markers}. All models of these fixtures will be made publicly available on our website.


\subsubsection{Object categories}
\label{app-sec-obj}
\begin{figure}[t!]
\vskip 0.2in
\begin{center}
\centerline{\includegraphics[width=\columnwidth]{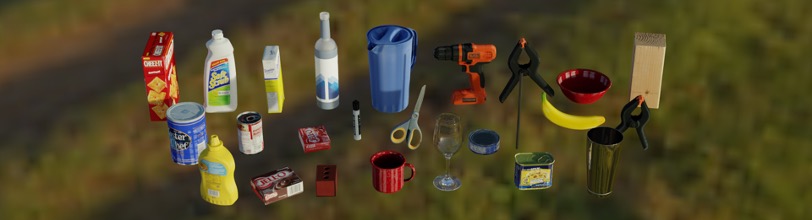}}
\caption{\textbf{Synthetic rendering of 25 object models in VinT-Sim.}}
\label{obj-sim}
\end{center}
\vskip -0.4in
\end{figure}

\begin{figure}[t!]
\vskip 0.2in
\begin{center}
\centerline{\includegraphics[width=\columnwidth]{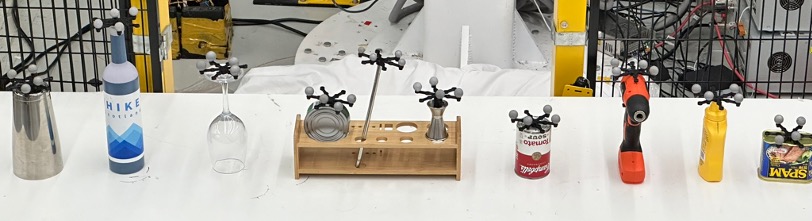}}
\caption{\textbf{Selected 10 objects used in the VinT-Real dataset.}}
\label{obj-real}
\end{center}
\vskip -0.4in
\end{figure}

In constructing the VinT-6D, we applied the following criteria for object selection: (1) \emph{Held in Hand}: Objects of an appropriate scale that can be comfortably held in hand. (2) \emph{Commonality}: Objects should be commonplace in everyday life and easily accessible.  (3) \emph{Variety}: The selection should cover a diverse range of materials, including plastics, metal, and glass. The field of robotic manipulation is increasingly focusing on domestic applications, such as household robots designed for tasks in laundry rooms or kitchens. These innovations are particularly significant for supporting independent living for the elderly, assisted living facilities, and addressing various healthcare challenges. In the VinT-Sim dataset, while we initially chose 21 objects from the Yale-CMU-Berkeley (YCB) dataset \cite{calli2017yale}, noted for their handheld size and frequent use in robotics research, we realized that some objects are difficult to find in stores, especially for those researchers not in US. Therefore, they are not enough for us to provide a generic object-in-hand benchmark for robotic perception researchers to design and test their algorithms. To enhance practicability, we incorporated 5 high-quality scanned objects from daily life, featuring transparent and reflective properties, as rendered in \cref{obj-sim}. For VinT-Real, we selected ten objects, such as a tomato soup can, power drill, and mustard bottle, detailed in \cref{obj-real}. These everyday items, selected for their specific functions, require particular grasp poses for tactile interaction. Additionally, objects like a metal shaker and wine glass vary in size and material distribution, with some, like the stir, presenting a challenge in fingertip touch during rotational holding to prevent dropping.

\begin{figure}[t!]
\vskip 0.2in
\begin{center}
\centerline{\includegraphics[width=1\columnwidth]{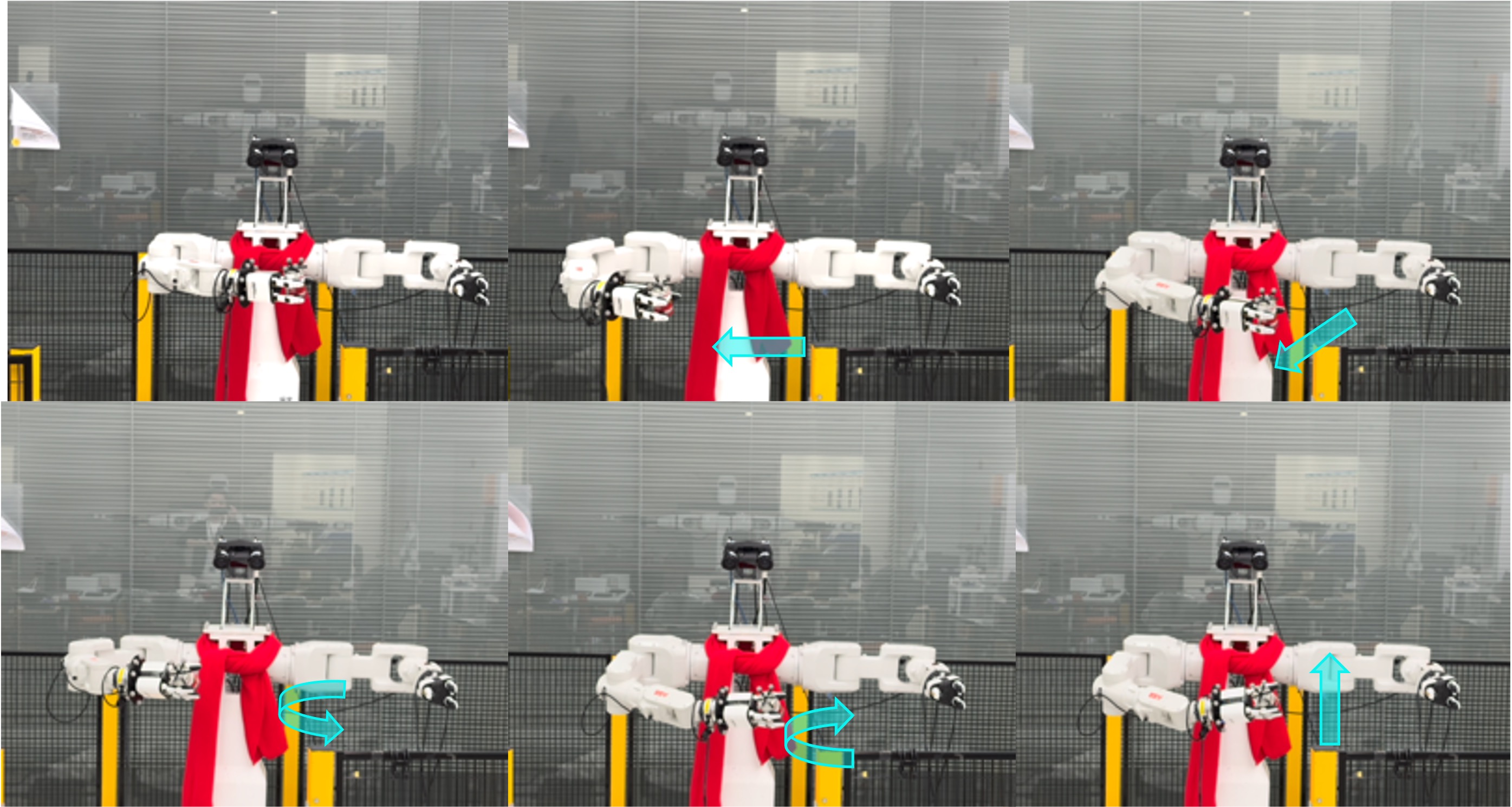}}
\caption{\textbf{Data collection mirroring toddler-like exploration.}}
\label{collect-toddlers.}
\end{center}
\vskip -0.2in
\end{figure}




\end{document}